\documentclass{article} 
\usepackage[preprint]{colm2026_conference}

\usepackage{microtype}
\usepackage{hyperref}
\usepackage{url}
\usepackage{booktabs}
\usepackage{amsmath}
\usepackage{graphicx}
\usepackage{wrapfig}
\usepackage{subcaption}
\usepackage{multirow}

\usepackage{lineno}

\definecolor{darkblue}{rgb}{0, 0, 0.5}
\hypersetup{colorlinks=true, citecolor=darkblue, linkcolor=darkblue, urlcolor=darkblue}

\usepackage{comment}

\title{Grid2Matrix: Revealing Digital Agnosia in Vision-Language Models}

\author{
    Yunkai Zhang\textsuperscript{1}*, 
    Linda Li\textsuperscript{2}, 
    Yingxin Cui\textsuperscript{2}, 
    Xiyuan Ruan\textsuperscript{2}, \\
    \textbf{~Zeyu Zheng\textsuperscript{1}, 
    Kezhen Chen\textsuperscript{3}, 
    Yi Zhang\textsuperscript{2}, 
    Diji Yang\textsuperscript{2}\thanks{Corresponding to: yunkai\_zhang@berkeley.edu, dyang39@ucsc.edu}}
    \\ \vspace{0.2cm}
    \textsuperscript{1}BAIR, UC Berkeley \quad
    \textsuperscript{2}UC Santa Cruz \quad
    \textsuperscript{3}Analogy AI\\
}

\begin{document}

\ifcolmsubmission
\linenumbers
\fi

\maketitle

\begin{abstract}
    Vision-Language Models (VLMs) excel on many multimodal reasoning benchmarks, but these evaluations often do not require an exhaustive readout of the image and can therefore obscure failures in faithfully capturing all visual details. 
    We introduce Grid2Matrix (G2M), a controlled benchmark in which a model is shown a color grid and a color-to-number mapping, and must output the corresponding matrix. By varying grid size and the number of colors, G2M provides a simple way to increase visual complexity while minimizing semantic confounds. We find that VLMs exhibit a sharp early collapse in zero-shot end-to-end evaluation, failing on surprisingly small grids rather than degrading gradually as the task becomes denser.
    We probe the visual encoders of VLMs from two representative families and find that they preserve substantially more of the grid information than the corresponding end-to-end outputs. This suggests that the failure is not explained by visual encoding alone, but also reflects a gap between what remains recoverable from visual features and what is ultimately expressed in language. We term this gap \textit{Digital Agnosia}.
    Further analyses show that these errors are highly structured and depend strongly on how grid cells overlap with visual patch boundaries. We also find that common strategies such as model scaling and multimodal alignment do not fully eliminate this failure mode.
    We expect G2M to serve as a useful testbed for understanding where and how VLMs lose fine visual details, and for evaluating tasks where missing even small visual details can matter, such as tables, charts, forms, and GUIs.
\end{abstract}

\section{Introduction}
Recent Vision-Language Models (VLMs) have achieved remarkable success on multimodal reasoning benchmarks, in some cases rivaling or surpassing human experts on complex tasks~\citep{alibaba2025qwen3vl,wang2025internvl3}. This progress is commonly attributed to a modular architecture that combines a powerful vision encoder with a large language model through a learned alignment interface~\citep{ghosh2024exploring}. As a result, modern VLMs can recognize high-level concepts, integrate visual evidence with world knowledge, and solve increasingly sophisticated multimodal problems~\citep{li2025survey}. However, many of these benchmarks do not require an exhaustive readout of the image: models can often succeed by relying on sparse salient cues, semantic regularities, or linguistic priors. This leaves open a basic question: how well do current VLMs handle tasks that require faithful dense spatial readout? Many practical multimodal tasks, including table parsing, chart reading, form understanding, and GUI interaction, depend precisely on this ability.

We ask whether a VLM can faithfully transcribe an image when success depends on reading all local details rather than on high-level interpretation. Our results suggest that often it cannot. Even a task as simple as converting a colored grid into a matrix exposes a severe and systematic failure mode.

To study this failure mode, we introduce Grid2Matrix (G2M), a diagnostic benchmark centered on dense spatial transcription. In G2M, the model is given an image of an $N\times N$ color grid together with a color-to-integer dictionary, and must output the corresponding matrix exactly. Because the inputs are synthetic and the task requires only direct visual transcription, G2M strips away the semantic confounds typical of natural images to focus strictly on dense spatial perception. At the same time, it provides a precise and continuous way to vary difficulty by controlling grid size and number of colors. This makes it possible to measure not only whether models fail, but also how and when that failure emerges as visual density increases.

Using G2M, we first show that frontier VLMs exhibit a sharp early collapse in zero-shot end-to-end performance. Across both proprietary and open-weight model families, Exact Match drops to zero on surprisingly small grids, often long before the underlying visual signal should be physically unrecoverable. This degradation is also abrupt rather than gradual: instead of steadily decaying with density, models often fail abruptly once the grid reaches a modest resolution. The result is a striking mismatch between strong performance on semantic multimodal benchmarks and weak performance on exhaustive spatial transcription.

To determine whether this collapse reflects a failure of visual encoding or a downstream failure of access and expression, we probe the isolated vision encoders of open-weight VLMs. These probes recover substantially more of the underlying grid structure than the corresponding end-to-end VLM outputs, even in regimes where the full model has already collapsed. This dissociation suggests that the failure cannot be attributed solely to absent visual information in the visual encoder. We term this representation-to-expression gap as ``Digital Agnosia'', by analogy to the neurological condition in which visual sensation is intact but recognition is impaired. Specifically, it refers to a failure mode in which visual information is encoded internally to a far greater extent than it is ultimately expressed through language output.

We then use G2M to characterize this failure mode in greater detail. Spatial heatmaps show that model errors are highly structured rather than random. Patch-grid alignment reveals that the recoverability of dense spatial detail depends critically on how grid cells interact with the encoder’s patch boundaries. Further analyses of model scaling and multimodal alignment show that neither larger models nor stronger aligned encoders straightforwardly eliminate the problem. Our main findings are: 
\begin{itemize}
    \item Frontier VLMs collapse on dense grids, often at surprisingly small sizes.
    \item Isolated vision encoders retain substantially more recoverable structure than end-to-end outputs reveal.
    \item The resulting failures are systematically shaped by spatial morphology, patch-grid alignment, model scaling, and multimodal alignment.
\end{itemize}
Overall, we position G2M as a controlled diagnostic benchmark for studying dense spatial fidelity and for analyzing how such failures emerge\footnote{Our code is available here: \url{https://github.com/zhykoties/Grid2Matrix_DigitalAgnosia}.}.

\section{The Grid2Matrix Benchmark}
\label{sec:benchmark_setup}

To study Digital Agnosia under controlled conditions, we introduce Grid2Matrix (G2M), a diagnostic benchmark for dense spatial transcription. Unlike benchmarks built on natural images, where models may rely on sparse salient cues, world knowledge, or linguistic priors, G2M uses procedurally generated color grids. This controlled design does not eliminate every non-perceptual factor, but it substantially reduces semantic confounds and provides a clean testbed for studying when visual structure is preserved, lost, or distorted.

\subsection{Task Formulation}
Each example consists of an image of an $N \times N$ grid in which every cell is filled with a solid color, together with a color-to-integer dictionary. The model must output the corresponding $N \times N$ integer matrix in text form, preserving the exact spatial arrangement of the grid. Because the ground-truth structure is known by construction, the task enables unambiguous evaluation at both the whole-grid and per-cell levels. For a human, this task requires little semantic inference and mainly tests precise visual tracking. For a VLM, it directly measures whether fine-grained spatial information can be extracted from the image and faithfully expressed through language.

\subsection{Parameterized Difficulty Gradient}
G2M varies difficulty along two axes while keeping the image resolution fixed at $512 \times 512$ pixels. The first is \textbf{grid size} ($N \times N$), which ranges from small layouts such as $3 \times 3$ to dense configurations up to $64 \times 64$. The second is \textbf{number of colors} ($C$), the number of unique colors in the grid, which we vary from $3$ to $10$. Increasing $N$ stresses spatial resolution, while increasing $C$ makes the information denser.

\begin{wrapfigure}{R}{0.5\textwidth}
    \vspace{-10pt} 
    \centering
    \begin{subfigure}[b]{0.45\linewidth}
        \centering
        \includegraphics[width=\linewidth]{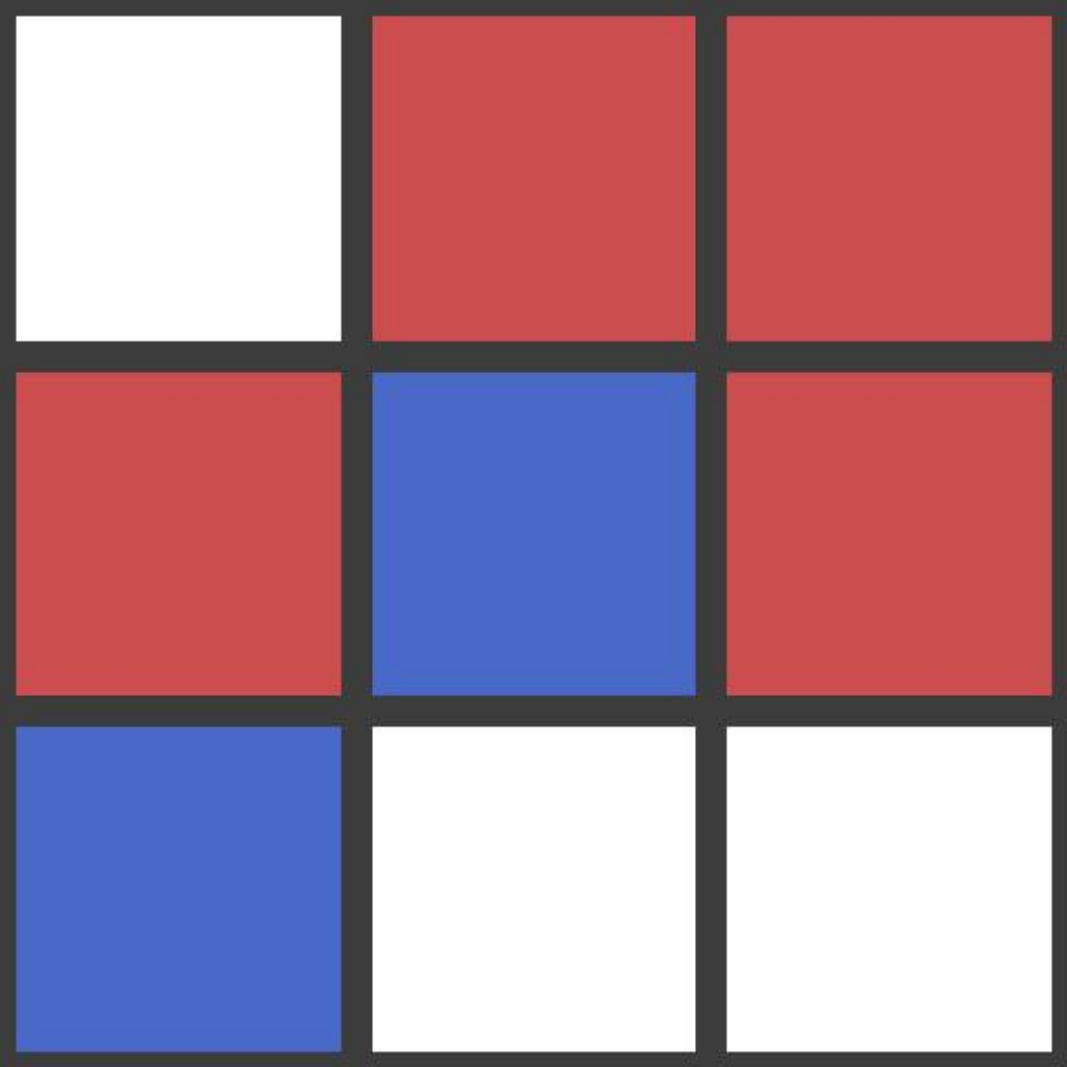} 
        \caption{$3 \times 3$}
        \label{fig:grid_3x3}
    \end{subfigure}
    \hfill
    \begin{subfigure}[b]{0.45\linewidth}
        \centering
        \includegraphics[width=\linewidth]{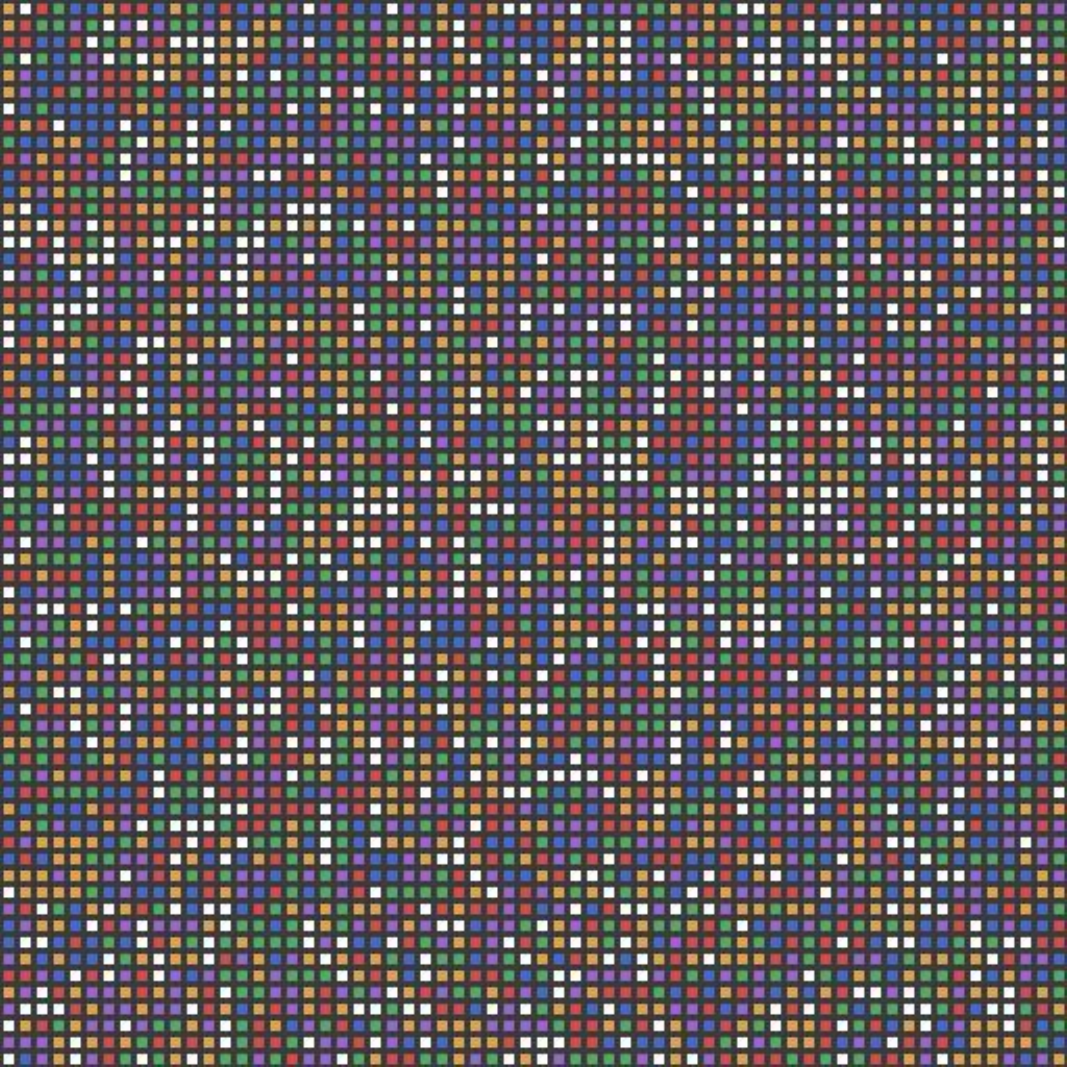} 
        \caption{$64 \times 64$}
        \label{fig:grid_64x64}
    \end{subfigure}
    \caption{Samples from G2M. Difficulty ranges from (a) simple tests to (b) dense settings that exceed standard patch resolution.}
    \label{fig:grid_samples}
    \vspace{-10pt} 
\end{wrapfigure}

For example, the $3 \times 3$ grid in Figure~\ref{fig:grid_samples}(a), together with the mapping $\{\text{White}: 0, \text{Red}: 1, \text{Blue}: 2\}$, corresponds to the matrix $[[0,1,1],[1,2,1],[2,0,0]]$. Figure~\ref{fig:grid_samples} visualizes the difficulty spectrum of the benchmark. While the low-resolution regime serves as a sanity check for instruction following, the high-resolution regime becomes a direct stress test for perceptual resolution. At high densities, the number of grid cells can exceed the number of visual patches, forcing the model to compress the distinct spatial location and color of multiple cells into a single patch embedding.

We report two complementary metrics. \textbf{Exact Match} is a binary grid-level score that requires the entire predicted matrix to match the ground truth exactly. \textbf{Cell Accuracy} is the fraction of cells predicted correctly. Exact Match captures whether a model can serialize the full structure without error, while Cell Accuracy supports finer-grained analysis of where and how failures emerge.

\subsection{Evaluation Scope and Probing Setup}
To test whether the bottleneck appears across representative model families, our zero-shot experiments\footnote{To reduce confounds from formatting and decoding variability, we use deterministic generation and a robust post-parser; details are in Appendix \ref{app:evaluation_details}.} include proprietary models from the GPT-5 series (GPT-5-mini, GPT-5.2) and Gemini-3 series (Gemini-3-Flash-Preview, Gemini-3-Pro-Preview), along with leading open-weight models from the InternVL3.5 and Qwen3-VL families. For the open-weight models, we further localize the source of failure by isolating the vision encoder (VE) before the multimodal projection layer and training a shallow convolutional spatial probe on its output representations to predict the grid layout. Comparing this probe-based readout with the final end-to-end textual output lets us distinguish information that is present in the visual representation from information that remains accessible after multimodal projection and language generation.

\section{Empirical Observations: Diagnosing Digital Agnosia}
\label{sec:results_diagnosis}

\subsection{End-to-End Evaluation: Early Collapse}
\label{sec:end_to_end}

We begin by evaluating the end-to-end zero-shot performance of frontier VLMs (GPT-5-mini and Gemini-3-Flash) on the G2M benchmark. Both models were tasked with transcribing grids of varying dimensions using a fixed dictionary of 3 colors.

\begin{wraptable}{l}{0.46\textwidth}
\vspace{-0.6em}
\centering
\small
\setlength{\tabcolsep}{5pt}
\renewcommand{\arraystretch}{1.05}
\caption{Zero-shot \textit{Exact / Cell} accuracies (\%) of proprietary models on 3-color grids.}
\label{tab:closed_source_zeroshot}
\begin{tabular}{@{}lcc@{}}
\toprule
\textbf{Grid} & \textbf{GPT-5-mini} & \textbf{Gemini-3-Flash} \\
\midrule
$3\times3$   & 100 / 100   & 100 / 100   \\
$6\times6$   & 97.3 / 99.9 & 99.0 / 100.0 \\
$9\times9$   & 0.0 / 65.6  & 97.3 / 100.0 \\
$12\times12$ & 0.0 / 41.5  & 69.7 / 99.2 \\
$20\times20$ & 0.0 / 33.7  & 0.0 / 87.8 \\
$32\times32$ & 0.0 / 0.0   & 0.0 / 38.5 \\
\bottomrule
\end{tabular}
\vspace{-0.8em}
\end{wraptable}

Table~\ref{tab:closed_source_zeroshot} shows that performance degrades sharply as spatial density increases. On simple $3 \times 3$ grids, both models achieve perfect Exact Match scores. As the grid size grows, however, accuracy drops quickly. Among proprietary models, Gemini-3-Flash demonstrates significantly higher resilience than GPT-5-mini. For instance, GPT-5-mini's Exact Match collapses to 0.00\% at just $9 \times 9$, whereas Gemini-3-flash-preview maintains a 97.33\% Exact Match at the same resolution. Even at $20 \times 20$, where neither model can produce an exact match, Gemini-3-Flash retains a Cell Accuracy of 87.77\%, compared with 33.73\% for GPT-5-mini. At $32 \times 32$, both models collapse toward the random baseline. Because the task uses three colors, random guessing yields about 33\% Cell Accuracy, indicating a complete loss of spatial structure.

\begin{wrapfigure}{r}{0.5\textwidth}
    \vspace{-10pt}
    \centering
    \includegraphics[width=\linewidth]{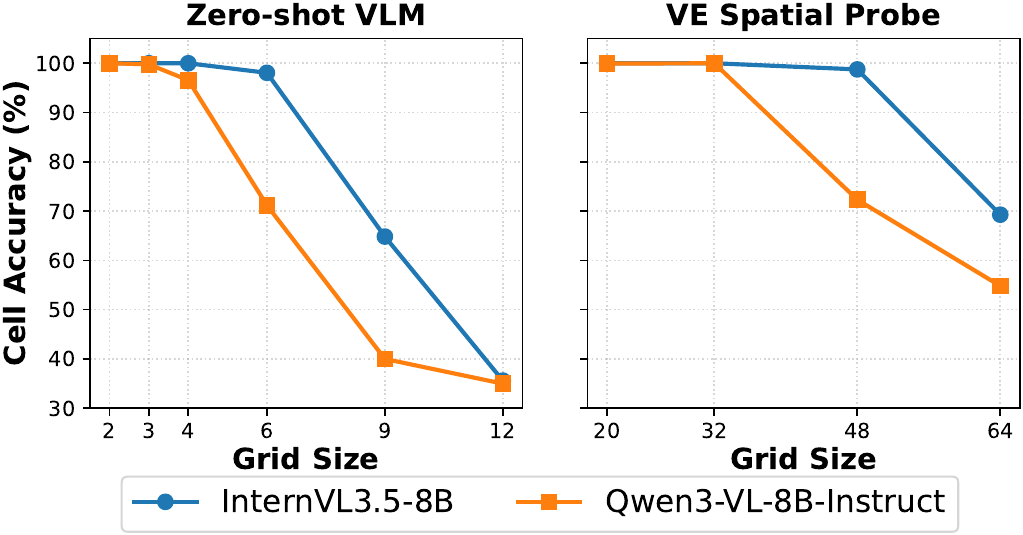} 
    \caption{Cell accuracy of open-weight models under zero-shot generation and VE probing. While zero-shot performance collapses rapidly as grid size increases, substantially higher accuracy remains recoverable from frozen VE features under supervised probing.}
    \label{fig:open-source-8b}
    \vspace{-10pt}
\end{wrapfigure}

The same pattern appears in the open-weight models in Figure~\ref{fig:open-source-8b} (left). Both InternVL3.5-8B and Qwen3-VL-8B-Instruct transcribe $2 \times 2$ grids perfectly, but performance deteriorates quickly as grid size increases. Qwen's Cell Accuracy has already dropped to 71.19\% by $4 \times 4$. InternVL maintains 62\% Exact Match at $6 \times 6$, but by $9 \times 9$, neither model can produce an exact match.

The systemic failure across model families on grids as small as $9 \times 9$ indicates a severe limitation in end-to-end dense spatial transcription. However, this evaluation treats the VLM as a black box and does not reveal whether the collapse is fully explained by missing spatial information in the vision encoder, or whether substantially more structure remains recoverable from visual features than is ultimately expressed by the full model. To investigate this question, we next isolate and probe the internal visual representations.

\subsection{Locating the Bottleneck: The Spatial Probe}
\label{sec:ve_probe}

To investigate this question, we probe the frozen vision encoders of the open-weight models using a shallow spatial readout model. Our goal is to test how much of the grid structure remains recoverable from visual features before multimodal compression and language generation. We extract representations from the vision encoders of InternVL3.5-8B and Qwen3-VL-8B-Instruct \textit{before} their respective multimodal compression layers (the Vision-Language Merger for Qwen and the Pixel Shuffle for InternVL). We then freeze these encoders and train a shallow convolutional probe on their final hidden states to predict the grid layout. This lets us test whether the vision encoder outputs still contain the information needed to complete the transcription before multimodal compression and language generation.

As shown in Figure~\ref{fig:open-source-8b}, the gap between the end-to-end VLM output and the VE probe is striking. In the zero-shot evaluation, models quickly degrade toward the random baseline of 33\% cell accuracy: for example, Qwen3-VL drops to 34.98\% cell accuracy at $12 \times 12$. In contrast, the Spatial Probe retains high accuracies well beyond this size. On a $32 \times 32$ grid, the probes for both InternVL3.5 and Qwen3-VL achieve 100.00\% Cell Accuracy.

The probing results also show that InternVL outperforms Qwen3-VL at extreme densities. InternVL maintains 98.75\% Cell Accuracy at $48 \times 48$, compared with 72.31\% for Qwen. This gap mirrors InternVL's advantage in the zero-shot evaluation. Importantly, we aim to keep it a fair comparison: at this stage of the architecture, both models process the input image into exactly 1024 patch tokens. In fact, InternVL achieves this superior structural retention despite using a slightly smaller hidden dimension (1024) than Qwen3-VL (1152). Even at $64 \times 64$, both probes remain well above the 33\% random-guessing baseline, indicating that substantial spatial structure is still recoverable from the extracted VE features.

Taken together, these findings provide strong diagnostic evidence: the VEs successfully capture and maintain the details of the grid. Instead, they point to a substantial bottleneck in how dense spatial information is preserved, accessed, or expressed after visual encoding. We refer to this representation-to-expression gap as \textit{Digital Agnosia}.

\section{Deep Dive: Spatial Morphology of Digital Agnosia}

This section characterizes the morphology of Digital Agnosia from three angles. First, we show that model errors are highly structured rather than random, revealing stable spatial failure patterns across architectures (Section~\ref{sec:spatial_morphology}). Second, we show that these failures depend strongly on patch-grid alignment, which affects how dense spatial detail is preserved or lost (Section~\ref{sec:patch_grid_geometry}). Third, we show that the number of colors affects probing and zero-shot generation differently, suggesting that dense visual transcription is shaped not only by what visual structure is recoverable, but also by how the structure is accessed and expressed downstream (Appendix~\ref{sec:color_ablations}).

\subsection{Visualizing the Error Landscape: Diagnostic Heatmaps}
\label{sec:spatial_morphology}

\begin{figure}[b]
    \centering
    \includegraphics[width=0.9\linewidth]{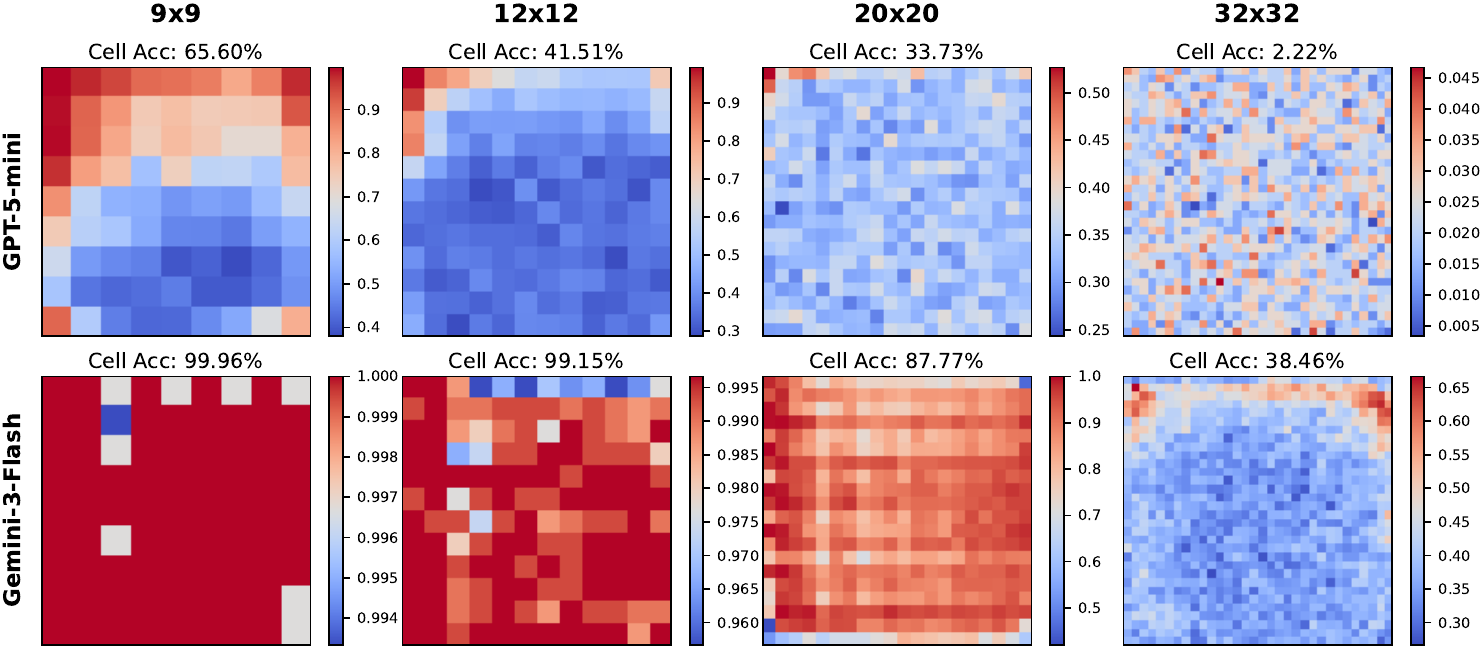} 
    \caption{Diagnostic spatial heatmaps for proprietary models for grid sizes from $9 \times 9$ to $32 \times 32$. Red indicates high cell accuracy and blue indicates lower accuracy.}
    \label{fig:diagnostic_heatmaps_close}
\end{figure}

We begin with the proprietary models in Figure~\ref{fig:diagnostic_heatmaps_close}. Their errors are clearly not uniform across the grid. GPT-5-mini shows a strong bias toward the top-left corner: on harder settings such as $12 \times 12$ and $20 \times 20$, accuracy remains relatively high only in that region, while the rest of the grid collapses.

Gemini-3-Flash shows a different pattern. Perhaps surprisingly, it frequently struggles with the top-most and bottom-most rows of the grid, which is particularly evident at the $20 \times 20$ resolution. By $32 \times 32$, the model can only maintain relatively higher accuracies around the top-left and top-right corners.

\begin{figure}[t]
    \centering
    \includegraphics[width=0.9\linewidth]{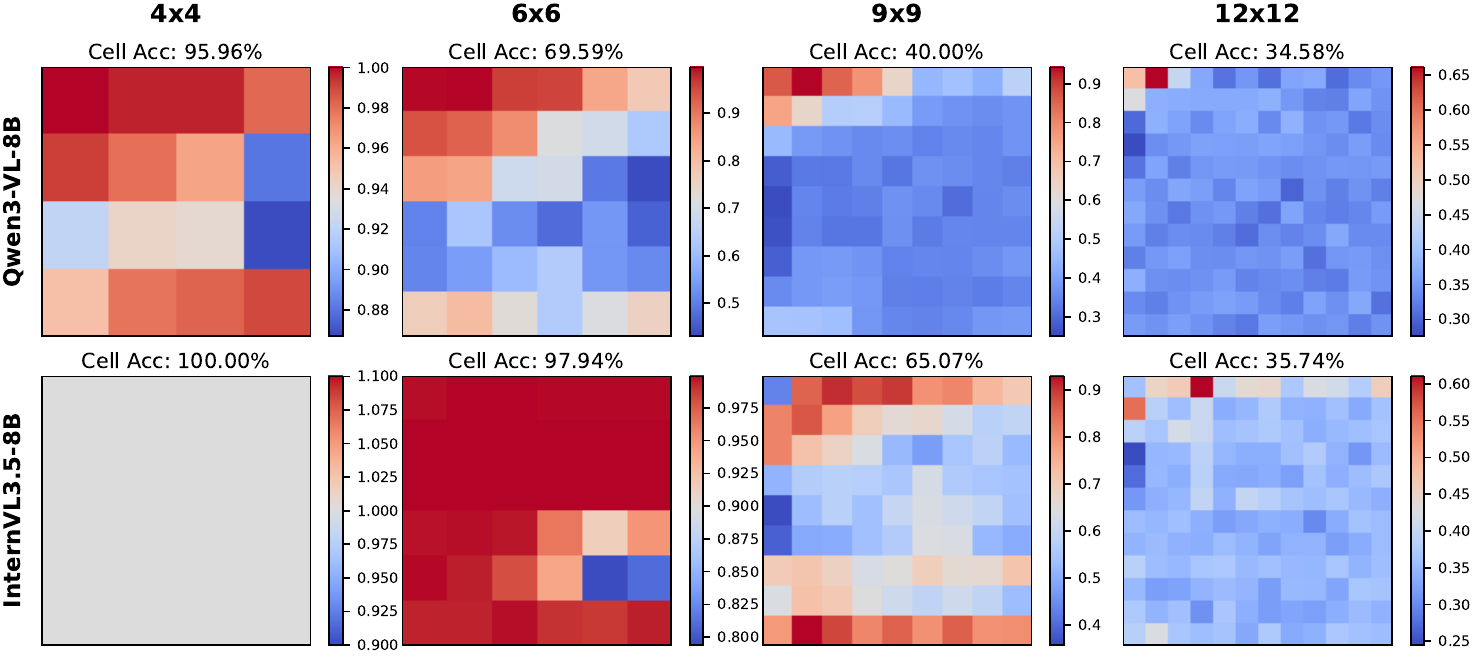} 
    \caption{Diagnostic spatial heatmaps for open-weight models for grid sizes from $4 \times 4$ to $12 \times 12$. Red indicates high cell accuracy and blue indicates lower accuracy.}
    \label{fig:diagnostic_heatmaps_open}
\end{figure}

The open-weight 8B models in Figure~\ref{fig:diagnostic_heatmaps_open} also show some structured patterns. In both InternVL3.5 and Qwen3-VL, accuracy generally declines as generation moves to the right and then downward through the grid, consistent with an autoregressive tracking failure. This suggests that the models struggle to maintain precise spatial state over long output sequences. Interestingly, the trend partially reverses near the very end of the output, where the last row or last few rows are often predicted somewhat more accurately.

\begin{figure}[ht]
    \centering
    \includegraphics[width=0.85\linewidth]{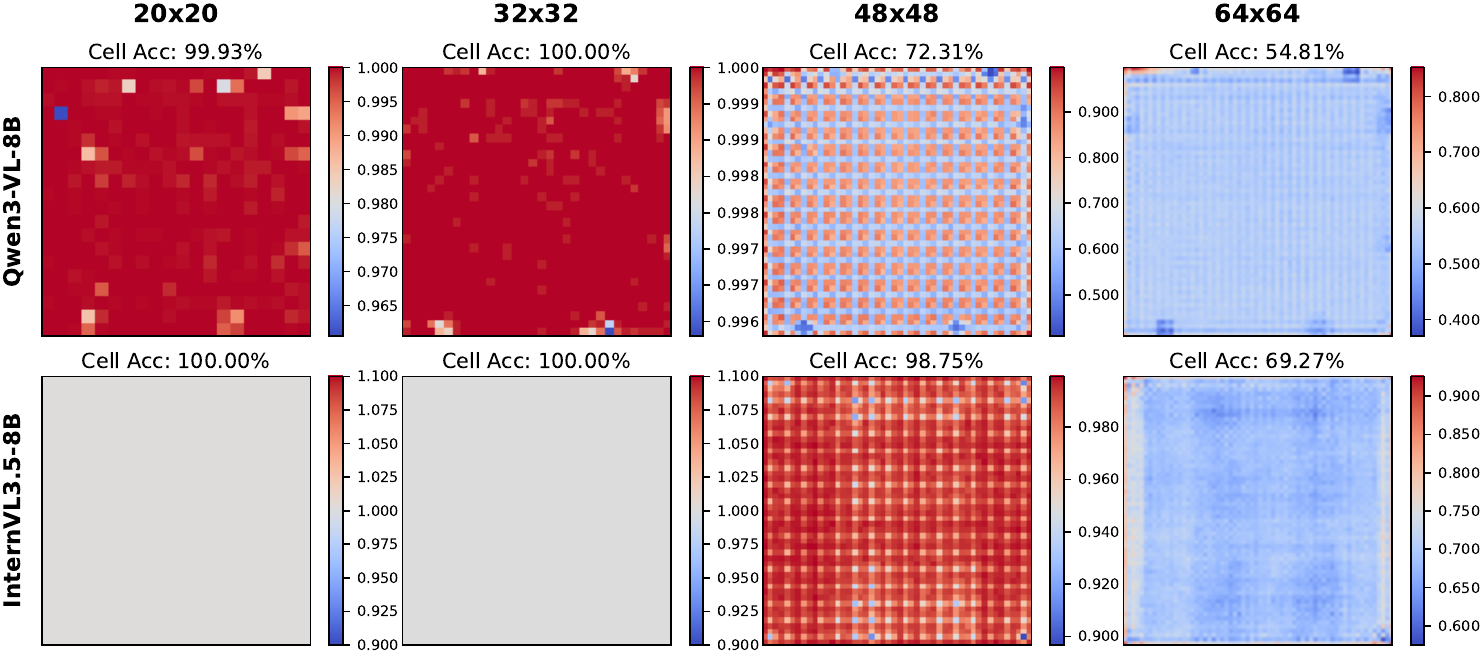} 
    \caption{Diagnostic spatial heatmaps for open-weight model VEs (via Spatial Probing) for grid sizes from $20 \times 20$ to $64 \times 64$. Red indicates high cell accuracy and blue indicates lower accuracy.}
    \label{fig:diagnostic_heatmaps_open_probing}
\end{figure}

Figure~\ref{fig:diagnostic_heatmaps_open_probing} applies the same analysis to the extracted vision encoders using the spatial probes from Section~\ref{sec:ve_probe}. Interestingly, the probe heatmaps show that the vision encoders themselves have stable spatial blind spots, which place an upper bound on how clearly the full VLM can represent the image. For example, the InternVL probe at $64 \times 64$ is more accurate near the borders and along the vertical middle line, with lower accuracy elsewhere. The Qwen3-VL probe shows a different pattern, with relatively high accuracy near the corners, alongside distinct clumps of low-accuracy cells distributed throughout the grid. Crucially, these blind spots are consistently located at the exact same positions regardless of the grid size, suggesting they are inherent artifacts of the VE.

\subsection{Grid-Patch Alignment}
\label{sec:patch_grid_geometry}

\begin{figure}[t]
    \centering
    \includegraphics[width=0.7\linewidth]{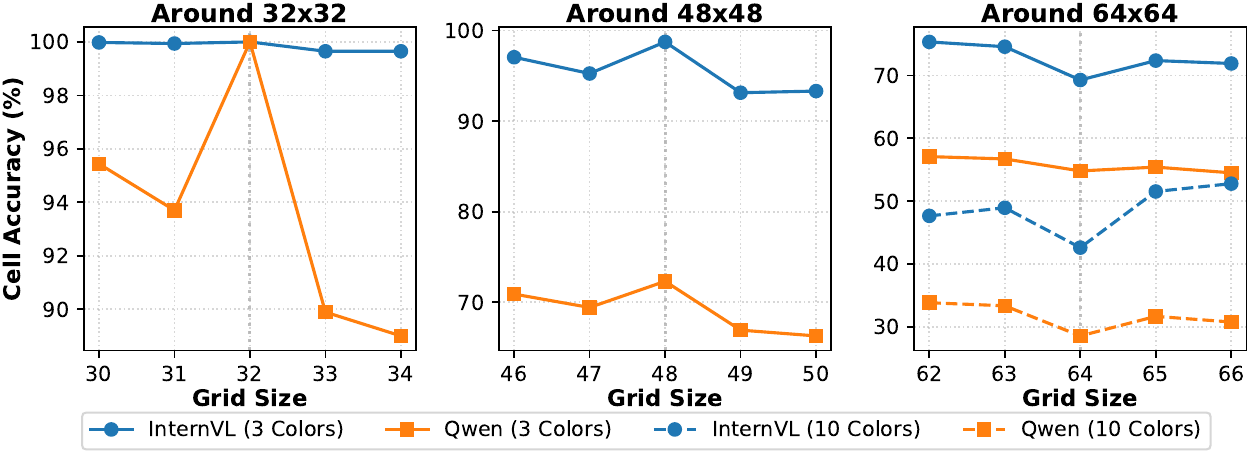} 
    \caption{Grid-patch alignment across varying grid sizes. Performance generally peaks at patch-aligned resolutions (e.g., $32 \times 32$, $48 \times 48$), but this trend breaks at $64 \times 64$, which underperforms its immediate misaligned neighbors.}
    \label{fig:aliasing_effect}
\end{figure}

We keep the input image fixed at $512 \times 512$ and evaluate frozen VE probes, so varying the grid size changes only how cells align with the vision encoder's $16 \times 16$ patch boundaries.

Figure~\ref{fig:aliasing_effect} shows that accuracy is not determined by grid density alone. As grids become denser, performance generally declines, but neighboring grid sizes can differ substantially depending on how their cells fall relative to the patch grid. This effect appears in both model families and in both the 3-color and 10-color settings.

The key factor is how each cell intersects patch boundaries. At a high level, a cell can be fully contained within a patch (``interior''), contained but touching a boundary (``edge''), or split across boundaries (``cross''). These cases are not equally recoverable: cells that are more fully contained within a patch tend to be recovered more accurately, while recoverability declines as patch-boundary conflict increases.

This helps explain the local peaks at $32 \times 32$ and $48 \times 48$. For a $512 \times 512$ image with $16 \times 16$ patches, a $32 \times 32$ grid gives exactly one cell per patch, while a $48 \times 48$ grid gives $3 \times 3$ cells in each $2 \times 2$ patch neighborhood (equivalently, $1.5 \times 1.5$ cells per patch).

The same framework also explains the $64 \times 64$ case, which might seem counterintuitive at first. Here, each $16 \times 16$ patch contains exactly $2 \times 2$ cells, so every cell falls to the ``edge'' category. By contrast, nearby misaligned grids contain a mixture of interaction types, including some ``interior'' cells that are easier to recover. Those easier cells can lift the overall average enough for a slightly misaligned neighbor to outperform $64 \times 64$. In this sense, the $64 \times 64$ result is a composition effect: aggregate accuracy depends on the distribution of interaction types, not just on whether the grid is globally aligned. 

Taken together, these results support a broader finding: dense spatial errors are strongly shaped by grid-patch alignment. They are therefore not random failures, and cannot be explained by grid density alone. Appendix~\ref{app:boundary_interaction_analysis} provides a more detailed analysis of the $64 \times 64$ case, including a full breakdown of boundary-interaction types.

\section{Scaling and Alignment Paradoxes}
\label{sec:scaling_paradoxes}

Can we overcome ``Digital Agnosia'' simply by leveraging standard industry practices, such as scaling model capacity or stronger alignment between VE and the LLM? In this section, we examine both factors separately. Our results show that although scaling and alignment can improve some aspects of performance, neither straightforwardly eliminates the representation-to-expression gap revealed by G2M.

\subsection{The Inefficacy of Model Scaling}
\label{sec:inefficacy_scaling}

\begin{wrapfigure}{R}{0.5\textwidth}
    \vspace{-10pt}
    \centering
    \includegraphics[width=\linewidth]{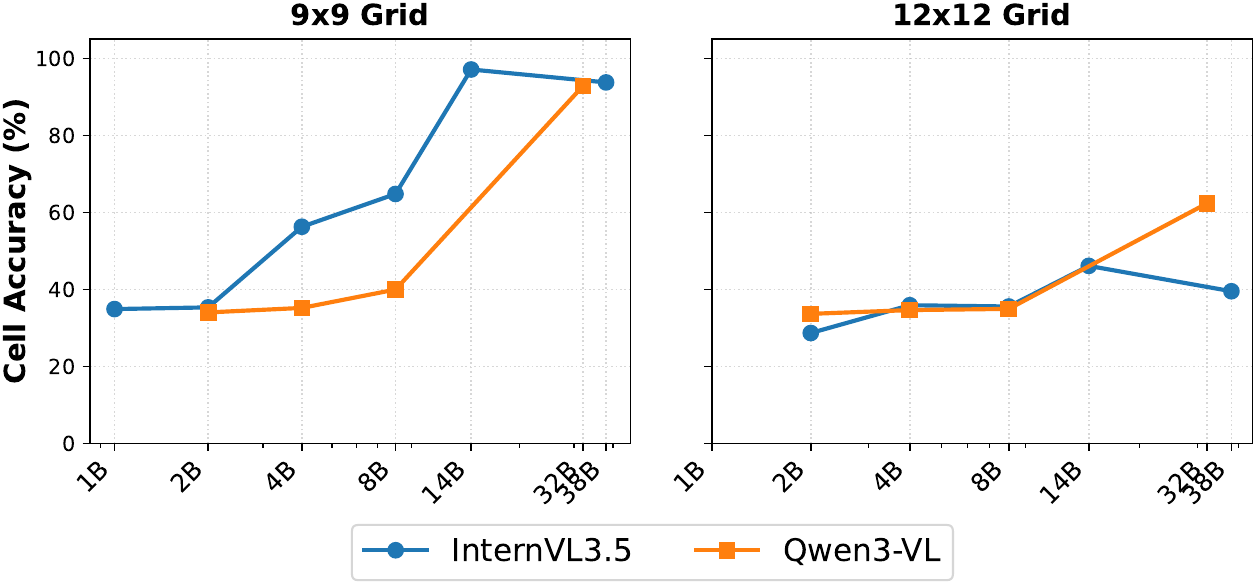} 
    \caption{Zero-shot scaling on open-weight models for $9 \times 9$ and $12 \times 12$ grids.}
    \label{fig:scaling_zeroshot_open}
    \vspace{-10pt}
\end{wrapfigure}

A natural assumption is that massively scaling model parameters might mitigate spatial collapse. To examine this, we separate the scaling of the VE from the scaling of the language-model backbone within the open-weight model families. As shown in Figure~\ref{fig:scaling_zeroshot_open} and Table~\ref{tab:scaling_probing}, scaling has different effects in Qwen3-VL and InternVL3.5, suggesting that model size alone does not uniformly translate into better dense spatial fidelity.

\begin{table}[htbp]
\centering
\caption{Spatial probe Cell Accuracy on dense grids. We only show the results for $48 \times 48$ and $64 \times 64$ here, since accuracies are saturated for smaller grids.}
\label{tab:scaling_probing}
\begin{tabular}{ll c cc}
\toprule
\multirow{2}{*}{\textbf{Model Family}} & \multirow{2}{*}{\textbf{VLM Size}} & \multirow{2}{*}{\textbf{Vision Encoder Backbone}} & \multicolumn{2}{c}{\textbf{Spatial Probe Accuracy}} \\
\cmidrule(lr){4-5}
& & & \textbf{$48 \times 48$} & \textbf{$64 \times 64$} \\
\midrule
\multirow{2}{*}{\textbf{Qwen3-VL}}  
& 2B / 4B & SigLIP2-Large (300M) & 81.59\% & 57.47\% \\
& 8B / 32B & SigLIP2-SO (400M)    & 72.31\% & 54.81\% \\
\midrule
\multirow{2}{*}{\textbf{InternVL3.5}}
& 1B--14B & InternViT (300M)     & 98.75\% & 69.27\% \\
& 38B      & InternViT (6B)       & 100.00\% & 73.34\% \\
\bottomrule
\end{tabular}
\end{table}

\textbf{Case 1: Qwen3-VL --- LLM Scaling Offsets a Weaker VE.} 
For Qwen3-VL, larger end-to-end models yield strictly better zero-shot accuracies (e.g., 32B drastically outperforms 8B). Spatial probing, however, reveals the opposite pattern in the VE: the smaller 300M encoder used in the 2B and 4B models retains higher spatial accuracy (81.59\%) than the larger 400M encoder used in the 8B and 32B models (72.31\%). The fact that zero-shot performance improves despite a degraded visual representation suggests that the larger LLM backbone may be compensating via stronger feature integration and language-driven completion, rather than truly ``seeing'' the grid more clearly. This interpretation is consistent with the autoregressive degradation pattern observed in Section~\ref{sec:spatial_morphology}, which indicates that end-to-end failures are shaped not only by what the VE preserves, but also by how spatial information is maintained during sequential readout.

\paragraph{Case 2: InternVL3.5 --- VE Scaling Outruns Zero-Shot Gains.}
InternVL3.5 shows a different pattern. Here, scaling the VE clearly improves probing performance: moving from the 300M InternViT to the 6B InternViT increases spatial probe accuracy from 98.75\% to 100.00\% at $48 \times 48$, and from 69.27\% to 73.34\% at $64 \times 64$. Yet these gains are not reflected in zero-shot end-to-end performance. In fact, zero-shot Cell Accuracy drops from 46.15\% to 39.56\% on $12 \times 12$ grids when scaling from 14B to 38B.

One possible explanation is that the LLM backbone fails to effectively extract topological details from the massively expanded high-dimensional visual feature space (from 1,024 in InternViT-300M to 3,200 in InternViT-6B). This interpretation is consistent with unexpected performance regressions observed in the InternVL3.5 technical report \citep{wang2025internvl3}. For example, for visual grounding, scaling from the 14B model (using a 300M encoder) to the 38B model (using a 6B encoder) causes overall accuracy on the RefCOCO benchmark suite~\citep{kazemzadeh2014referitgame, mao2016generation} to drop from 94.7\% to 91.8\%. This regression is even more pronounced in GUI navigation tasks: accuracy on the ScreenSpot benchmark~\citep{cheng2024seeclick} falls sharply from 87.5\% to 81.0\%, and success rates on WebArena-Lite-v2~\citep{wang2025mmbench} drop from 12.3\% to 7.1\%. This regression is even more striking because, in the InternVL family, performance generally improves monotonically as model size increases from 1B to 14B; the 38B model is the point where this trend breaks. More broadly, this suggests that scaling the VE alone is not enough: while a scaled VE provides massive projection bandwidth, the LLM backbone and its alignment data must scale much more aggressively to fully utilize those dense visual features. As a direct consequence, Qwen3-VL-32B (62.37\%) significantly outperforms InternVL3.5-38B (39.56\%) on the $12 \times 12$ benchmark.

Taken together, these results show that scaling does not remove the gap in any simple way. In Qwen, the gains appear to come mainly from the language-model side, despite weaker VE recoverability. In InternVL, the VE becomes stronger, but the downstream model does not fully capitalize on it. To understand why these families behave differently, we next examine how their blind spots change across scale and alignment.

\subsection{Inherited Blind Spots and the Alignment Paradox}
\label{sec:alignment_paradox}

\begin{figure}[b]
    \centering
    \includegraphics[width=0.85\linewidth]{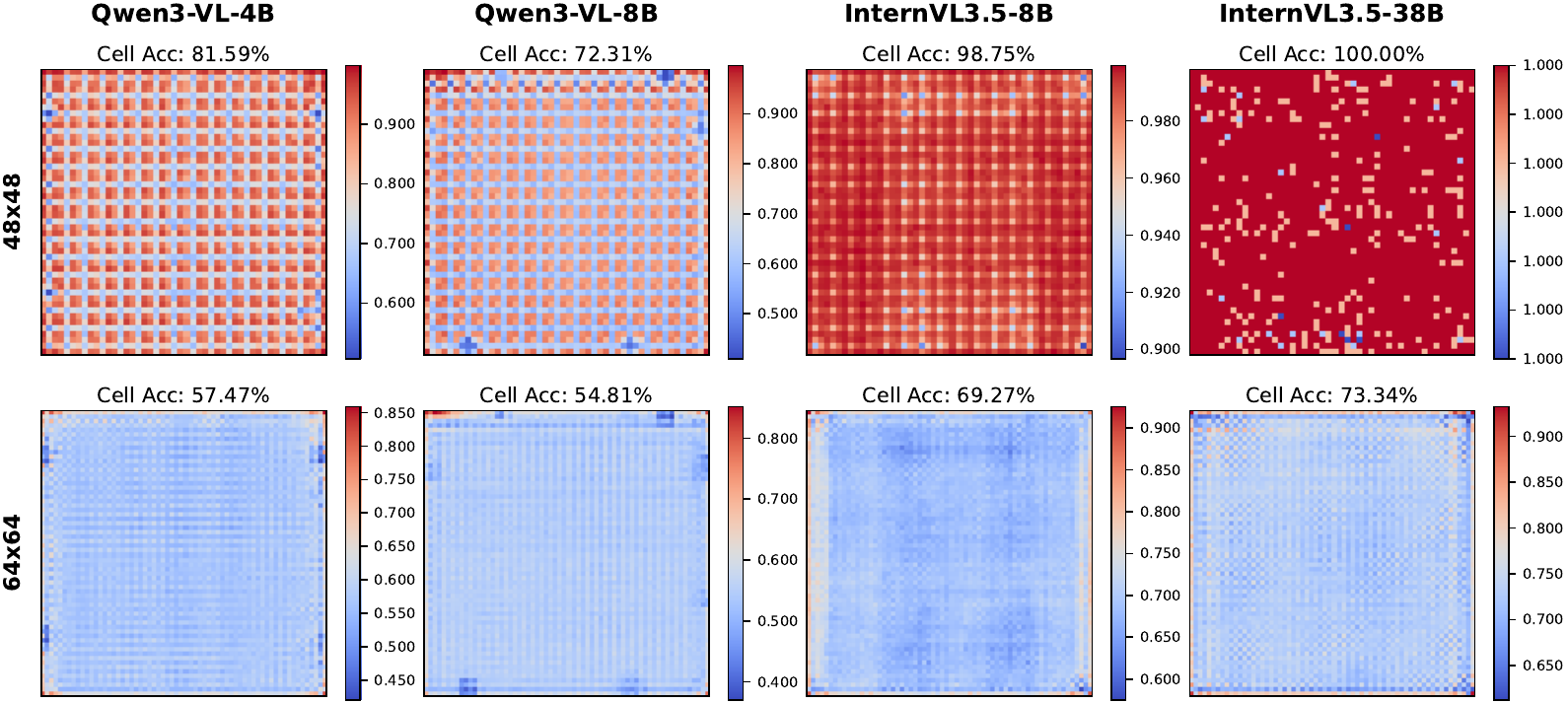} 
    \caption{Error heatmaps from spatial probes on open-weight models at $48 \times 48$ and $64 \times 64$.}
    \label{fig:heatmap_scale_open}
\end{figure}

Figure~\ref{fig:heatmap_scale_open} shows that blind spots evolve differently with scale in the two model families.

In Qwen3-VL, the main blind spots remain highly consistent across scales. Because the 300M and 400M SigLIP2 backbones share the same ViT architectures and pretraining data~\citep{tschannen2025siglip2, alibaba2025qwen3vl}, their shared errors likely reflect stable biases inherited from pretraining. Surprisingly, the Shape-Optimized (SO) 400M variant actually underperforms the 300M model by introducing new blind spots, specifically the two additional blind spots at the top edge and two at the bottom edge of the grid. One possible explanation is that this model was optimized for a $14 \times 14$ patch regime~\citep{alabdulmohsin2023getting, tschannen2025siglip2}, then adapted to a $16 \times 16$ version to match Qwen's $2 \times 2$ patch-merging projector. That retrofit may weaken fine-grained spatial reasoning. More broadly, the persistence of these blind spots suggests that once such spatial biases are present in the vision encoder, they may be difficult to remove later through scaling or alignment. This makes G2M especially valuable as an early diagnostic tool: it exposes these failures at the vision-encoder level, where they arise most directly and before they become harder to disentangle downstream.

Conversely, the differing error patterns in InternVLs suggest a potential spatial cost associated with knowledge distillation. At $64 \times 64$, the 300M model is relatively accurate along the vertical edges but weaker in the middle, whereas the 6B teacher model does not exhibit this central decay. Unlike SigLIP2, the 300M InternViT is explicitly distilled from the 6B teacher~\citep{chen2024internvl, opengvlab2024miniinternvl}. We hypothesize that while cross-architecture distillation effectively transfers global semantic alignment, it inadvertently discards the teacher's high-fidelity spatial routing. Knowledge distillation for vision encoders generally minimizes feature divergence on natural images, which may heavily prioritize high-level object semantics over pixel-perfect, dense spatial perception. Furthermore, this performance gap could also stem from the vast difference in raw model capacity: the 6B model's massive parameter count likely provides the necessary representational depth to resolve sub-patch details and suppress attention interference without relying on structurally flawed approximations. 

This contrast with Qwen suggests a more specific question: are the differing blind spots in InternVL introduced during multimodal alignment, or do they already originate in the underlying vanilla vision encoders? To test this directly, we compare \textit{vanilla} and \textit{extracted} vision encoders. Here, \textit{vanilla} refers to the original pretrained vision encoder before multimodal instruction tuning, while \textit{extracted} refers to the vision encoder taken from the fully aligned VLM. 

\begin{figure}[t]
    \centering
    \includegraphics[width=0.9\linewidth]{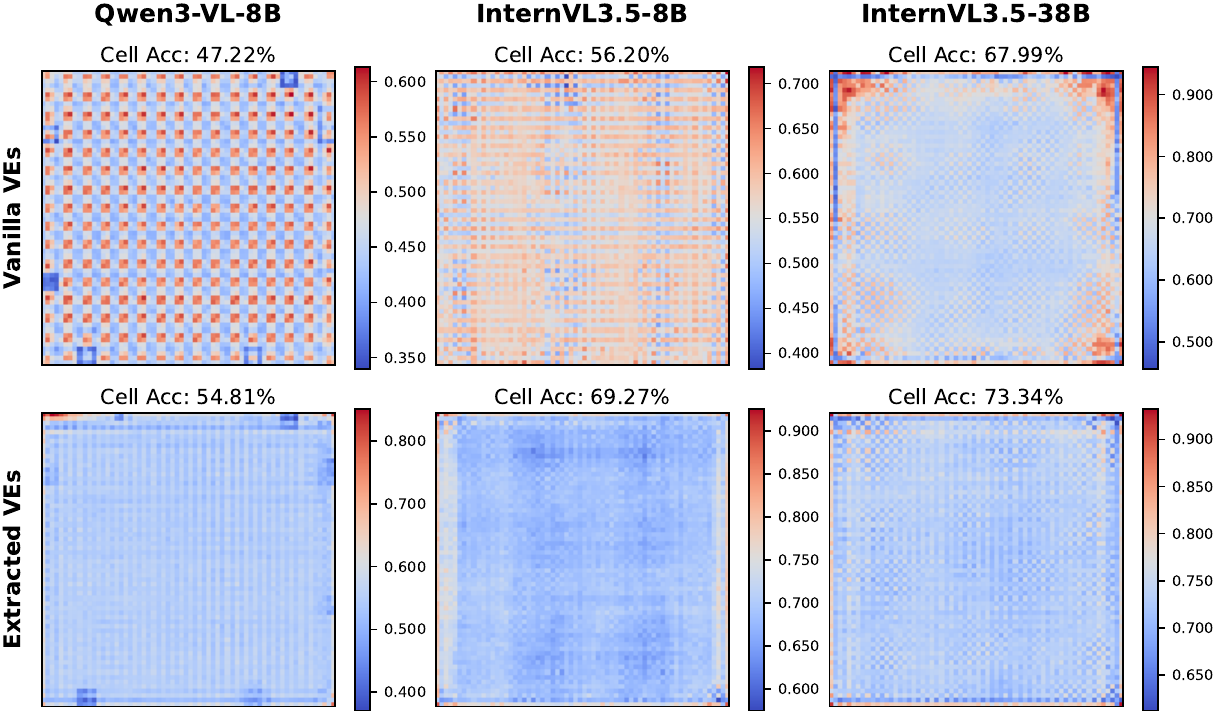} 
    \caption{Error heatmaps comparing vanilla and extracted vision encoders at $64 \times 64$. Alignment improves overall accuracy, but many of the base encoder's blind spots remain visible.}
    \label{fig:vanilla_vs_extracted}
\end{figure}

Figure~\ref{fig:vanilla_vs_extracted} shows that alignment improves performance across much of the grid. Overall accuracy increases, several low-accuracy regions become less severe, and some blind spots even mostly disappear. However, many blind spots remain. In Qwen, several low-accuracy regions in the extracted encoder remain in locations similar to those observed in the vanilla SigLIP2-SO, suggesting that these blind spots are deeply inherited and difficult to remove completely. In InternVL, by contrast, the vanilla 300M and 6B encoders already exhibit different blind-spot layouts, which helps explain why their extracted counterparts also differ. In other words, alignment improves spatial recoverability on top of the biases already present in the base encoder; it does not start from a neutral representation.

\begin{table}[ht]
\centering
\caption{Spatial probe Cell Accuracy on dense grids. Comparing vanilla and extracted vision encoders shows that multimodal alignment substantially improves spatial recoverability under probing.}
\label{tab:vanilla_vs_extracted}
\begin{tabular}{ll cccc}
\toprule
\multirow{2}{*}{\textbf{Model Family}} & \multirow{2}{*}{\textbf{Vision Encoder}} & \multicolumn{4}{c}{\textbf{Spatial Probe Accuracy}} \\
\cmidrule(lr){3-6}
& & \textbf{$20 \times 20$} & \textbf{$32 \times 32$} & \textbf{$48 \times 48$} & \textbf{$64 \times 64$} \\
\midrule
\multirow{2}{*}{\textbf{Qwen3-VL-8B}}
& SigLIP2-SO (400M) & 95.49\% & 75.39\% & 57.94\% & 47.23\% \\
& Extracted & \textbf{99.93\%} & \textbf{100.00\%} & \textbf{72.31\%} & \textbf{54.81\%} \\
\midrule
\multirow{2}{*}{\textbf{InternVL3.5-8B}}
& InternViT (300M)  & \textbf{100.00\%} & \textbf{100.00\%} & 70.89\% & 56.20\% \\
& Extracted & \textbf{100.00\%} & \textbf{100.00\%} & \textbf{98.75\%} & \textbf{69.27\%} \\
\midrule
\multirow{2}{*}{\textbf{InternVL3.5-38B}}
& InternViT (6B)    & \textbf{100.00\%} & \textbf{100.00\%} & 99.97\% & 67.99\% \\
& Extracted & \textbf{100.00\%} & \textbf{100.00\%} & \textbf{100.00\%} & \textbf{73.34\%} \\
\bottomrule
\end{tabular}
\end{table}

Table~\ref{tab:vanilla_vs_extracted} quantifies the improvement. Under spatial probing, extracted encoders substantially outperform their vanilla counterparts in all three settings, confirming that multimodal alignment can greatly improve spatial recoverability in the vision encoder.

Two factors may contribute to this improvement during the VLM alignment phase:
\begin{enumerate}
    \item \textbf{Architectural Retrofitting:} Foundational models like SigLIP often rely on absolute positional embeddings interpolated from smaller resolutions, which can result in ambiguous positional information. During alignment, models like Qwen3-VL discard these in favor of 2D Rotary Positional Embeddings (2D-RoPE), which better preserve relative spatial layout~\citep{alibaba2025qwen3vl}.
    \item \textbf{Grid-Centric Fine-Tuning:} During multimodal instruction-tuning, the vision encoder is unfrozen and exposed to document-level data, including charts, tables, and OCR tasks. This forces the encoder to specialize in parsing rigid row and column structures. This is corroborated by the Qwen3-VL technical report, which notes that fine-tuning the vision encoder yields significant performance gains on structured data~\citep{alibaba2025qwen3vl}.
\end{enumerate}

Taken together, these results lead to what we term the \textbf{Alignment Paradox}. Multimodal alignment improves how much dense spatial structure is recoverable from the vision encoder under probing, yet the same architecture can still fail to express that structure through zero-shot language generation. In other words, alignment can produce a stronger visual representation without fully correcting inherited blind spots or ensuring that the downstream language model will make full use of it. This reinforces the value of G2M as an early diagnostic benchmark.

\section{Related Work}

\subsection{Vision Language Models}
The contemporary landscape of Vision-Language Models (VLMs) is dominated by a modular architectural paradigm that synergizes visual perception with linguistic reasoning~\citep{li2025survey}. At a high level, these systems consist of three components: a pretrained VE responsible for extracting spatial and semantic features, a Large Language Model (LLM) backbone responsible for instruction following and reasoning, and an alignment module that maps visual representations into the LLM's input token space. While the foundational architecture was popularized by LLaVA \citep{liu2023visual}, which demonstrated the efficacy of simple linear projection, the field has since split into distinct research lineages aiming to scale this capability. The Qwen-VL series~\citep{alibaba2025qwen3vl} represents a multimodal design optimized for native-resolution processing and interleaved multimodal inputs, innovating in dynamic resolution processing and interleaved inputs to handle variable image aspect ratios natively. Conversely, the InternVL series \citep{wang2025internvl3} adopts a strong vision philosophy, scaling the parameter count of the vision encoder itself to rival the LLM in size, prioritizing high-fidelity feature extraction. In this work, we specifically select representative models from these two families: Qwen3-VL and InternVL3.5, to probe the perceptual limits of both design philosophies.

\subsection{Benchmarks for VLMs}
The current evaluation landscape for VLMs is predominantly defined by benchmarks that stress-test high-level semantic reasoning (\textit{thinking}) rather than low-level spatial fidelity (\textit{seeing}). \citet{li2025survey} conclude that the field mostly targets complex downstream capabilities, such as Visual Math Reasoning (e.g., MathVista~\citep{lu2023mathvista}, MM-Vet~\citep{yu2023mm}), Chart and Diagram Understanding (e.g., ChartQA~\citep{masry2022chartqa}, AI2D~\citep{kembhavi2016diagram}), or Multimodal General Intelligence (e.g., MMStar~\citep{chen2024we}, MMMU~\citep{yue2024mmmu}). These benchmarks evaluate the model's ability to synthesize visual information with world knowledge to solve complex reasoning tasks. While benchmarks focused on grounding exist, such as OCR tasks \citep{liu2024ocrbench} or hallucination detection (e.g., POPE \citep{li2023evaluating}), they largely test the retrieval of specific semantic entities (text or objects).

However, both high-level reasoning and entity-detection benchmarks share a common limitation: they allow models to succeed via sparse perception. In these tasks, the model needs only attend to salient regions or key visual tokens to derive the correct answer, effectively ignoring the global spatial structure and fine-grained details of the remaining image. Consequently, the field implicitly relies on the VE's pre-training objectives (e.g., masked reconstruction loss~\citep{he2022masked}) to achieve holistic perceptual acuity.

Recent work has begun to show that this assumption is problematic, confirming that critical visual information is heavily compromised during the VLM alignment process. Notably, \citet{fu2025hidden} demonstrate that VLMs frequently fail to use high-level visual representations, such as depth estimation and visual correspondence, that are already successfully captured by their isolated VEs. While that work exposes an alignment bottleneck for complex visual attributes, it operates entirely within the realm of semantic interpretation. A complementary gap remains at the level of dense spatial transcription: there are still few diagnostic benchmarks that directly penalize the loss of fine-grained but structurally important visual information. Our work addresses this gap by introducing a task in which success depends on preserving the full spatial structure of the image rather than attending only to salient regions.

\subsection{Perceptual Resolution}
Perceptual resolution characterizes the effective information transfer across the modality gap, distinct from raw input resolution or encoder patch size. It represents the finest granularity of visual detail that a Vision-Language Model can successfully preserve and map to linguistic tokens. This metric highlights a critical distinction in VLM architectures: the difference between the information capacity of the visual encoder and the actual spatial fidelity accessible to the language model.

This distinction is grounded in the cognitive science concept of visual agnosia \citep{farah2004visual}. In neurology, visual agnosia refers to a dissociation between \textit{sensation} (intact visual acuity and scanning) and \textit{perception} (the ability to recognize or describe the visual input) \citep{goodale2013sight}. In such cases, the sensory organs and primary cortex capture the image data correctly, but the downstream association cortex fails to synthesize it into a meaningful percept. Analogous to this biological condition, we define Digital Agnosia as the computational failure mode in VLMs where high-fidelity spatial details recoverable from the vision encoder are only partially preserved or expressed in downstream language output. This results in a state where the model's \textit{sensation} (encoder embedding) is intact, yet its \textit{perception} (LLM understanding) remains effectively blind to fine-grained structure.

\section{Conclusion}

We introduced Grid2Matrix (G2M), a controlled benchmark for testing whether Vision-Language Models can faithfully transcribe dense spatial structure when semantic confounds are minimized. Across both proprietary and open-weight models, we find a sharp early collapse in end-to-end performance on this simple visual task. Probing results suggest that this failure is not fully explained by missing visual information alone: isolated vision encoders retain substantially more recoverable grid structure than the corresponding end-to-end VLM outputs reveal. We use the term \textit{Digital Agnosia} to describe this representation-to-expression gap, in which visual information is encoded internally to a much greater extent than it is ultimately expressed through language output.

Beyond this core dissociation, G2M shows that the failure is highly structured rather than random. Its severity depends on factors such as grid-patch alignment, model scaling, and multimodal alignment, suggesting that dense spatial perception depends not only on what visual information is captured, but also on whether that information can be preserved, accessed, and expressed throughout the multimodal stack. Although G2M is synthetic and does not aim to capture the full distribution of natural multimodal tasks, we view it as a controlled stress test for a capability that matters in practice in layout-sensitive settings such as tables, charts, forms, and GUIs. We hope this work encourages the community to treat dense spatial perception as a first-class evaluation target, and to judge models not only by high-level multimodal reasoning, but also by their ability to retain and communicate fine-grained visual structure.

\section*{Acknowledgments}
Special thanks to Baochen Sun for his insightful discussions and constructive feedback on the early draft. We also thank Analogy AI for their support throughout this project.

\bibliography{colm2026_conference}
\bibliographystyle{colm2026_conference}

\newpage

\appendix

\section{Boundary-Interaction Analysis for the \texorpdfstring{$64 \times 64$}{64x64} Case}
\label{app:boundary_interaction_analysis}

\begin{figure}[t]
    \centering
    \includegraphics[width=0.85\linewidth]{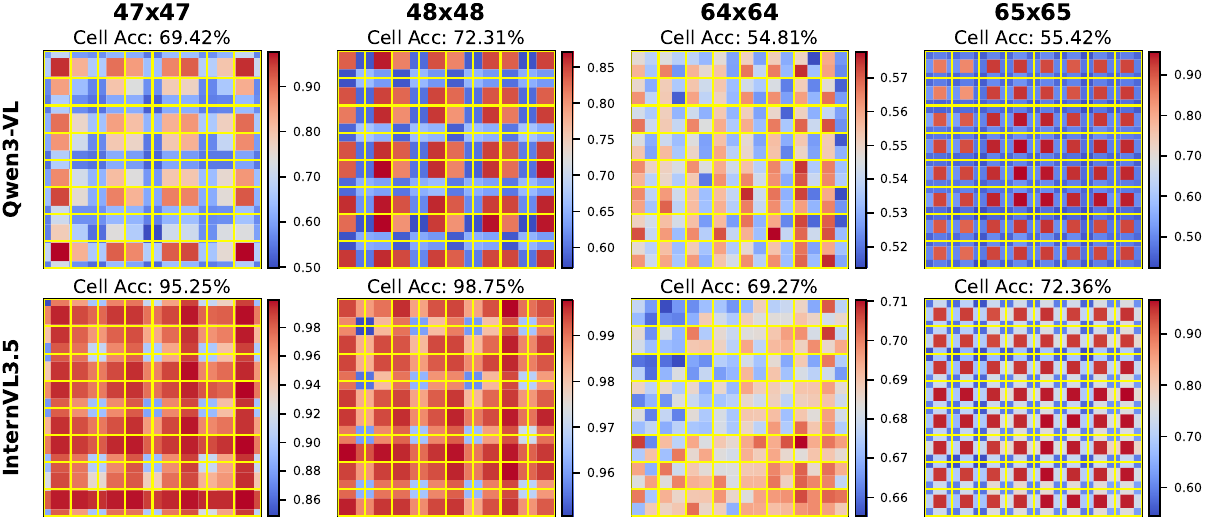} 
    \caption{Error heatmaps for $47 \times 47$, $48 \times 48$, $64 \times 64$, and $65 \times 65$. Yellow lines show the patch boundaries. For better visibility, we show only the center $8 \times 8$ patches. See Figures~\ref{fig:full-alias-internvl} and~\ref{fig:full-alias-qwen} for the full heatmaps.}
    \label{fig:aliasing_heatmap}
\end{figure}

\begin{figure}[t]
    \centering
    \includegraphics[width=\linewidth]{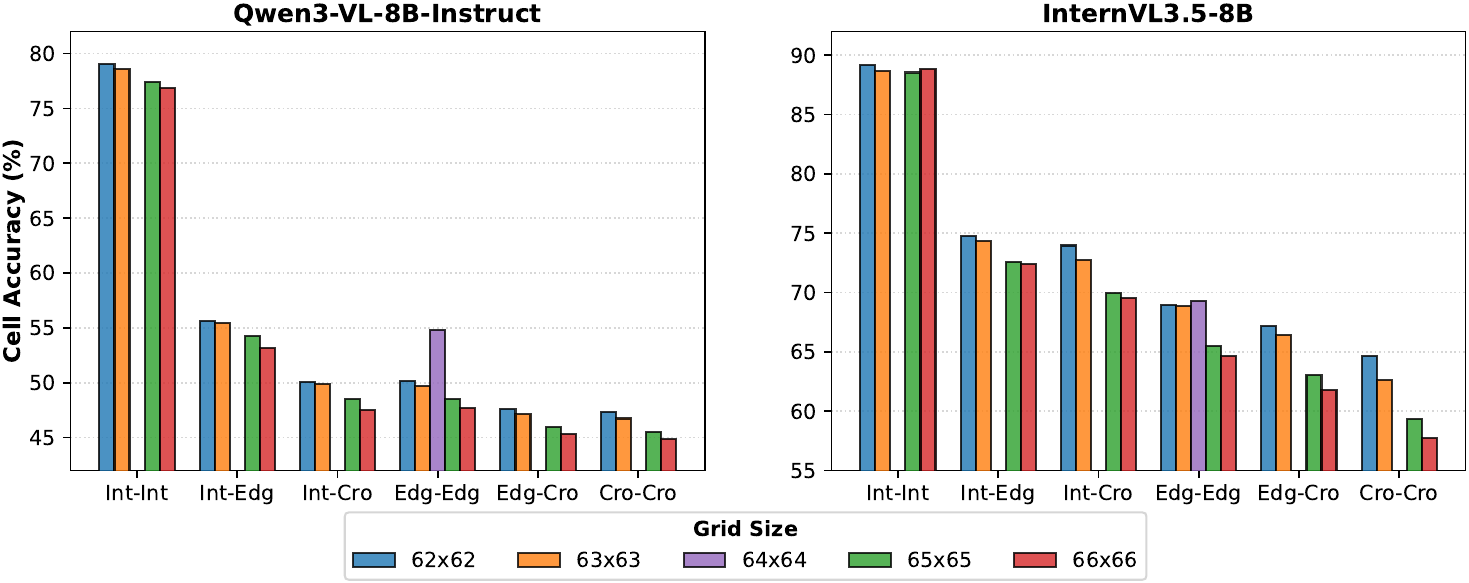}
    \caption{Cell accuracy across six boundary-interaction types near the $64 \times 64$ grid size. Because a $64 \times 64$ grid yields exactly $2 \times 2$ cells in each patch, 100\% of its cells fall into the Edg-Edg configuration. Nearby misaligned grids distribute their cells across all categories, revealing a strict degradation in accuracy as cells cross more patch boundaries.}
    \label{fig:aliasing_scenarios}
\end{figure}

In Section~\ref{sec:patch_grid_geometry}, we argued that dense spatial recoverability depends on how grid cells align with the vision encoder's patch boundaries. We now examine the $64 \times 64$ case in more detail to show that its behavior can be explained by the same idea.

A natural first intuition is that patch-aligned grid sizes should always perform best. This is true for settings such as $32 \times 32$ and $48 \times 48$, which form local peaks in Figure~\ref{fig:aliasing_effect}. For a $512 \times 512$ image with $16 \times 16$ patches, a $32 \times 32$ grid gives exactly one cell per patch, so each patch contains only one color and can represent that cell cleanly. Moving slightly away from this alignment causes cell boundaries to straddle neighboring patches, forcing the model to aggregate conflicting color signals. A similar local recovery appears at $48 \times 48$, where the grid yields exactly $3 \times 3$ cells in each $2 \times 2$ patch neighborhood. However, $64 \times 64$ underperforms some nearby grid sizes (e.g., $62 \times 62$ or $65 \times 65$). Rather than treating this as a separate phenomenon, we ask which boundary-interaction types appear most often at this grid size.

To make this concrete, we categorize each cell according to how it intersects patch boundaries along each axis. At a high level, a cell can be: (1) fully contained within a patch without touching a boundary (\textbf{Interior}), (2) fully contained but touching a boundary (\textbf{Edge}), or (3) split by a boundary (\textbf{Cross}). Combining the two axes yields six symmetric interaction types: Int-Int, Int-Edg, Int-Cro, Edg-Edg, Edg-Cro, and Cro-Cro.

This taxonomy helps explain Figures~\ref{fig:aliasing_heatmap} and~\ref{fig:aliasing_scenarios}. The zoomed heatmaps reveal a simple rule of area dominance: cells occupying more of a patch tend to be recovered more accurately, while cells with less patch area tend to be recovered less accurately. The interaction-type breakdown shows the same phenomenon quantitatively. Across nearby grid sizes, performance follows a staircase-like ordering: cells that are more fully contained within a patch tend to be recovered most accurately, while cells that touch or cross more boundaries become progressively harder to recover. In other words, recoverability declines as patch-boundary conflict increases.

Within this framework, the $64 \times 64$ case becomes easier to interpret. Because each $16 \times 16$ patch contains exactly $2 \times 2$ cells, every cell falls into the same Edg-Edg configuration: no cell lies safely in the interior of a patch, but no cell falls into the poorly-performed boundary-crossing cases either. By contrast, nearby misaligned grids contain a mixture of interaction types. Although some cells are harder to recover, those grids also include more favorable Int-Int or Int-Edg configurations, which can raise the overall average enough for a slightly misaligned neighbor to outperform $64 \times 64$. This is the composition effect referred to in Section~\ref{sec:patch_grid_geometry}: aggregate accuracy depends on the distribution of boundary-interaction types, not just on whether the grid is globally aligned. More broadly, the $64 \times 64$ result reinforces the same conclusion as the main text: dense spatial fidelity is governed not simply by grid density or global alignment, but by which cell-patch interaction types dominate the image.

\section{Color Ablations: Semantic Interference and Sensitivity}
\label{sec:color_ablations}

In this section, we expand from our default 3 colors (White, Red, Blue) up to 10 distinct colors to analyze how the number of colors interacts with spatial perception. By evaluating these permutations through both spatial probing (vision encoder only) and zero-shot generation (end-to-end VLM), we uncover a clear contrast between the raw feature extraction of the vision encoder and the final output generated by the language model.

\begin{table}[htbp]
\centering
\caption{Spatial Probing Cell Accuracy (\%) of the frozen VE across varying color diversities. The notation \textbf{3C} to \textbf{10C} denotes the number of distinct colors present in the grid.}
\label{tab:opensource_probing_color_ablation}
\resizebox{\textwidth}{!}{
\begin{tabular}{l cccccc cccccc}
\toprule
\multirow{2}{*}{\textbf{Grid Size}} & \multicolumn{6}{c}{\textbf{InternVL3.5-8B (Probe VE)}} & \multicolumn{6}{c}{\textbf{Qwen3-VL-8B-Instruct (Probe VE)}} \\
\cmidrule(lr){2-7} \cmidrule(lr){8-13}
& \textbf{3C} & \textbf{4C} & \textbf{5C} & \textbf{6C} & \textbf{8C} & \textbf{10C} & \textbf{3C} & \textbf{4C} & \textbf{5C} & \textbf{6C} & \textbf{8C} & \textbf{10C} \\
\midrule
\textbf{20 $\times$ 20} & 100.00 & 100.00 & 100.00 & 100.00 & 100.00 & 100.00 & 99.93  & 99.91  & 99.84  & 99.76  & 99.55  & 99.49  \\
\textbf{32 $\times$ 32} & 100.00 & 100.00 & 100.00 & 100.00 & 100.00 & 100.00 & 100.00 & 100.00 & 99.99  & 99.99  & 99.96  & 99.97  \\
\textbf{48 $\times$ 48} & 98.75  & 97.60  & 96.68  & 95.80  & 92.88  & 90.73  & 72.31  & 67.80  & 64.12  & 61.70  & 56.36  & 54.40  \\
\textbf{64 $\times$ 64} & 69.27  & 62.23  & 56.63  & 52.46  & 46.45  & 42.55  & 54.81  & 47.41  & 41.97  & 38.08  & 32.31  & 28.59  \\
\bottomrule
\end{tabular}
}
\end{table}

\begin{figure}[t]
    \centering
    \includegraphics[width=\linewidth]{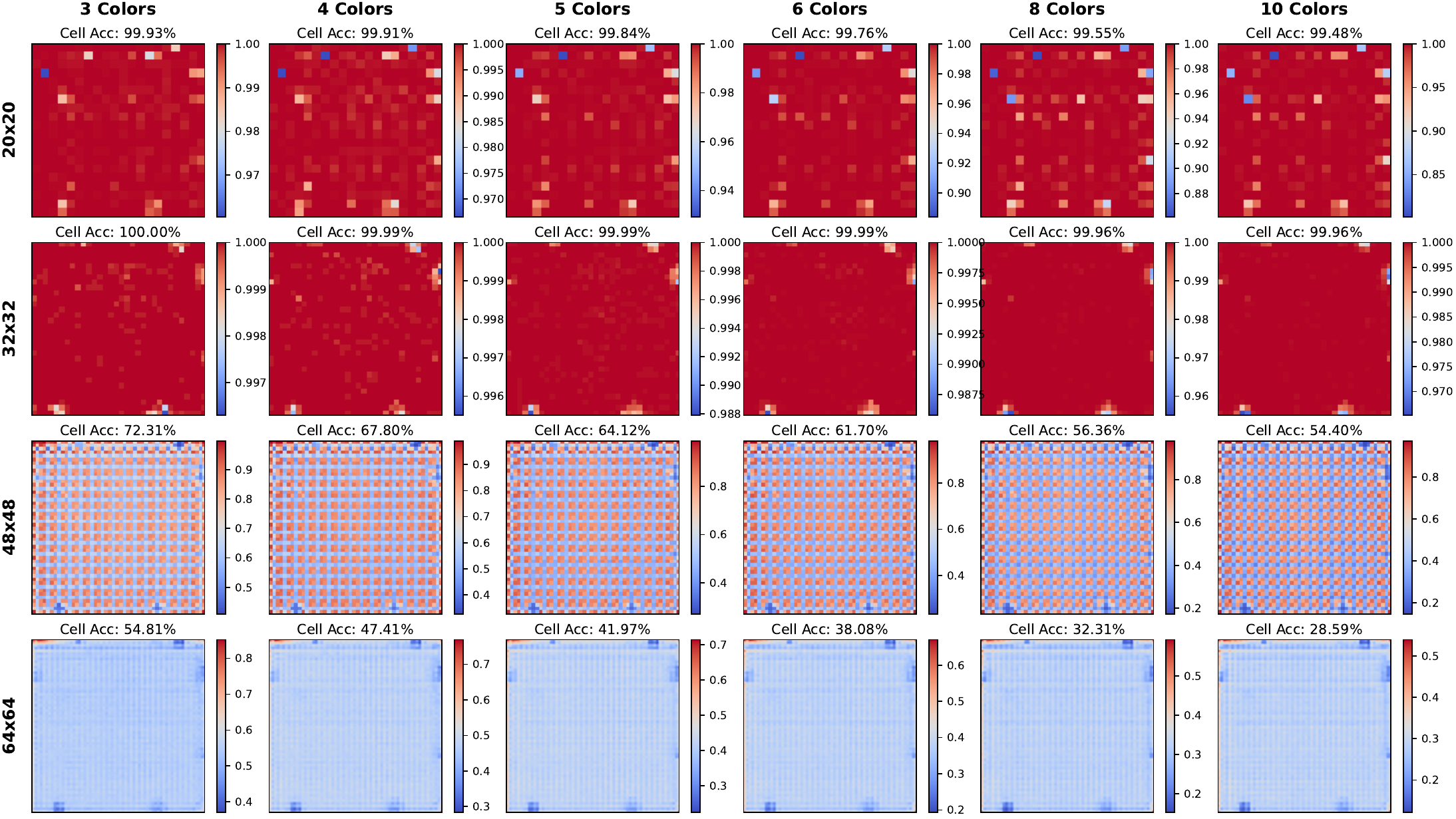} 
    \caption{VE spatial probing heatmaps for Qwen3-VL-8B-Instruct. While the blind spots persist across different color counts (see Appendix~\ref{sec:color_ablations}), the overarching performance patterns fluctuate across grid sizes due to the cells being divided by patch boundaries and area dominance effects analyzed in Appendix~\ref{sec:patch_grid_geometry}.}
    \label{fig:color_size_qwen_probing}
\end{figure}

\begin{figure}[t]
    \centering
    \includegraphics[width=\linewidth]{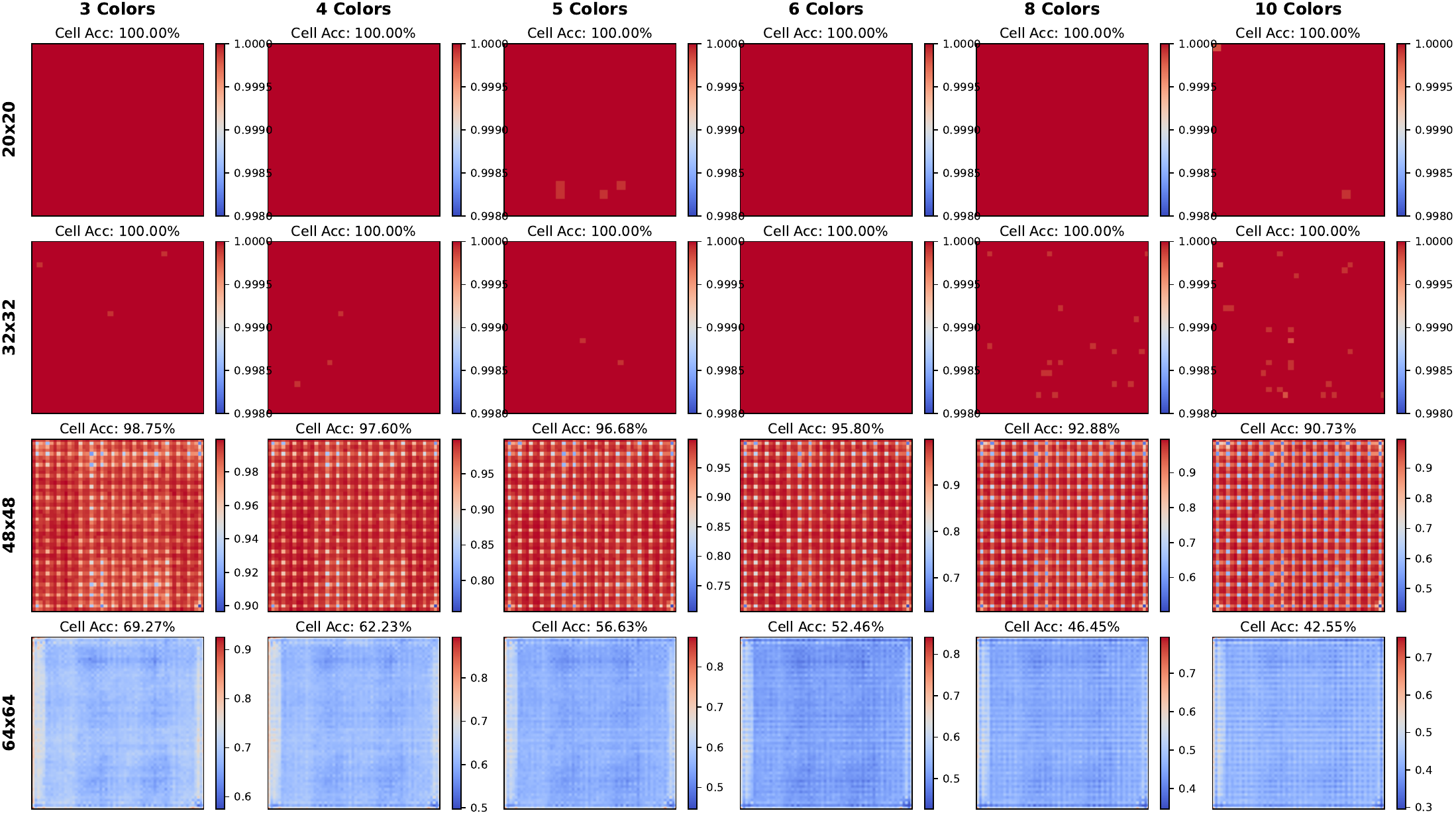} 
    \caption{VE spatial probing heatmaps for InternVL3.5-8B. While the blind spots persist across different color counts (see Appendix~\ref{sec:color_ablations}), the overarching performance patterns fluctuate across grid sizes due to the cells being divided by patch boundaries and area dominance effects analyzed in Appendix~\ref{sec:patch_grid_geometry}.}
    \label{fig:color_size_internvl_probing}
\end{figure}

\textbf{Probing Degrades with More Colors.} Under isolated spatial probing, increasing the number of colors strictly decreases Cell Accuracy across all grid sizes. The results are presented in Table~\ref{tab:opensource_probing_color_ablation}. As the palette expands, the visual features extracted by the vision encoder become increasingly blurry, struggling to resolve boundaries between diverse adjacent colors. However, heatmaps (Figure~\ref{fig:color_size_qwen_probing} and \ref{fig:color_size_internvl_probing}) reveal that the locations of the vision encoder's blind spots remain identical regardless of the number of colors. This provides strong empirical evidence that these perceptual blind spots are structural, architectural biases, completely independent of the color distribution, and persisting across varying levels of color diversity.

\textbf{Zero-Shot Improves with More Colors.} 
As shown in Table~\ref{tab:opensource_color_ablation}, evaluating the zero-shot, end-to-end VLM reveals a counter-intuitive phenomenon. While one might expect more colors to complicate the perceptual task, an expanded palette frequently leads to \textit{higher} zero-shot cell accuracies on moderately sized grids (e.g., $4 \times 4$ and $6 \times 6$ for Qwen, and $6 \times 6$ for InternVL). 

Furthermore, the difficulty of the task naturally increases with the number of colors: a random-guessing baseline achieves only 10\% Cell Accuracy in the 10-color setting, compared to 33.3\% with 3 colors. Despite this, on a $12 \times 12$ grid, both InternVL and Qwen score well above 10\% for 10 colors. Comparatively, their relative advantage over the random baseline is notably smaller in the 3-color setting. This demonstrates that the relative advantage of color diversity persists even as global spatial tracking begins to break down on larger grids.

\begin{table}[htbp]
\centering
\caption{Zero-Shot Cell Accuracy (\%) of Open-Weight VLMs across varying color diversities. The notation \textbf{3C} to \textbf{10C} denotes the number of distinct colors present in the grid.}
\label{tab:opensource_color_ablation}
\resizebox{\textwidth}{!}{
\begin{tabular}{l cccccc cccccc}
\toprule
\multirow{2}{*}{\textbf{Grid Size}} & \multicolumn{6}{c}{\textbf{InternVL3.5-8B}} & \multicolumn{6}{c}{\textbf{Qwen3-VL-8B-Instruct}} \\
\cmidrule(lr){2-7} \cmidrule(lr){8-13}
& \textbf{3C} & \textbf{4C} & \textbf{5C} & \textbf{6C} & \textbf{8C} & \textbf{10C} & \textbf{3C} & \textbf{4C} & \textbf{5C} & \textbf{6C} & \textbf{8C} & \textbf{10C} \\
\midrule
\textbf{3 $\times$ 3}   & 100.00 & 100.00 & 100.00 & 100.00 & 100.00 & 99.33 & 99.78 & 99.78 & 99.78 & 100.00 & 100.00 & 100.00 \\
\textbf{4 $\times$ 4}   & 100.00 & 99.84  & 100.00 & 99.99  & 100.00 & 99.99 & 95.86 & 98.30 & 98.88 & 99.35  & 99.69  & 99.78  \\
\textbf{6 $\times$ 6}   & 97.94  & 97.90  & 98.02  & 98.59  & 99.06  & 99.10 & 69.59 & 74.69 & 78.22 & 80.89  & 81.76  & 82.22  \\
\textbf{9 $\times$ 9}   & 65.07  & 62.07  & 63.14  & 64.47  & 65.36  & 63.14 & 40.00 & 38.59 & 37.34 & 36.98  & 35.20  & 34.03  \\
\textbf{12 $\times$ 12} & 35.74  & 28.43  & 24.55  & 22.13  & 19.94  & 17.39 & 34.58 & 26.53 & 22.62 & 19.39  & 16.42  & 14.67  \\
\bottomrule
\end{tabular}
}
\end{table}

We hypothesize that the zero-shot improvement with more colors reflects a retrieval problem during sequential readout rather than improved low-level perception. When there are fewer colors, many patches share highly similar visual embeddings, so the model must repeatedly pick out the correct local signal from many near-duplicates. This may cause attention to diffuse across similar candidates rather than stay anchored to a single location, making coordinate tracking harder. This intuition is consistent with findings from long-context retrieval benchmarks, where performance degrades when the relevant signal is less distinctive relative to distractors or harder to match directly~\citep{liu2024lost,modarressi2025nolima}.

A second factor is autoregressive instability. Standard transformers are prone to degeneration when forced to generate long repetitive sequences~\citep{Holtzman2020The}. In our setting, a small color palette produces long stretches of visually and symbolically repetitive structure, which may make the model more likely to lose its place or fall into repetitive decoding patterns. By contrast, a larger palette creates more distinctive local patterns, which may act as semantic landmarks that help the model maintain its position in the grid during sequential readout. This interpretation is also consistent with multimodal long-context benchmarks showing that LMMs are sensitive to visual distractors and to the placement of relevant information within the context~\citep{wu2025visual}.

This interpretation also aligns with recent concurrent work showing that VLMs experience catastrophic spatial localization failures on binary grids when cells lack distinct textual or semantic identities~\citep{levental2026squares}. Taken together, these results suggest that current VLMs rely heavily on distinctiveness and semantic separability when reading out dense spatial structure, rather than on a robust native coordinate mapping alone. We conjecture that zero-shot performance inversely correlates with the average contiguous color-blob size in the input grid, an empirical analysis we leave for future work. Finally, we observe that Qwen is substantially more sensitive to these color shifts than InternVL, particularly on $6 \times 6$ grids when comparing the 3-color and 10-color settings.

\begin{figure}[t]
    \centering
    \includegraphics[width=\linewidth]{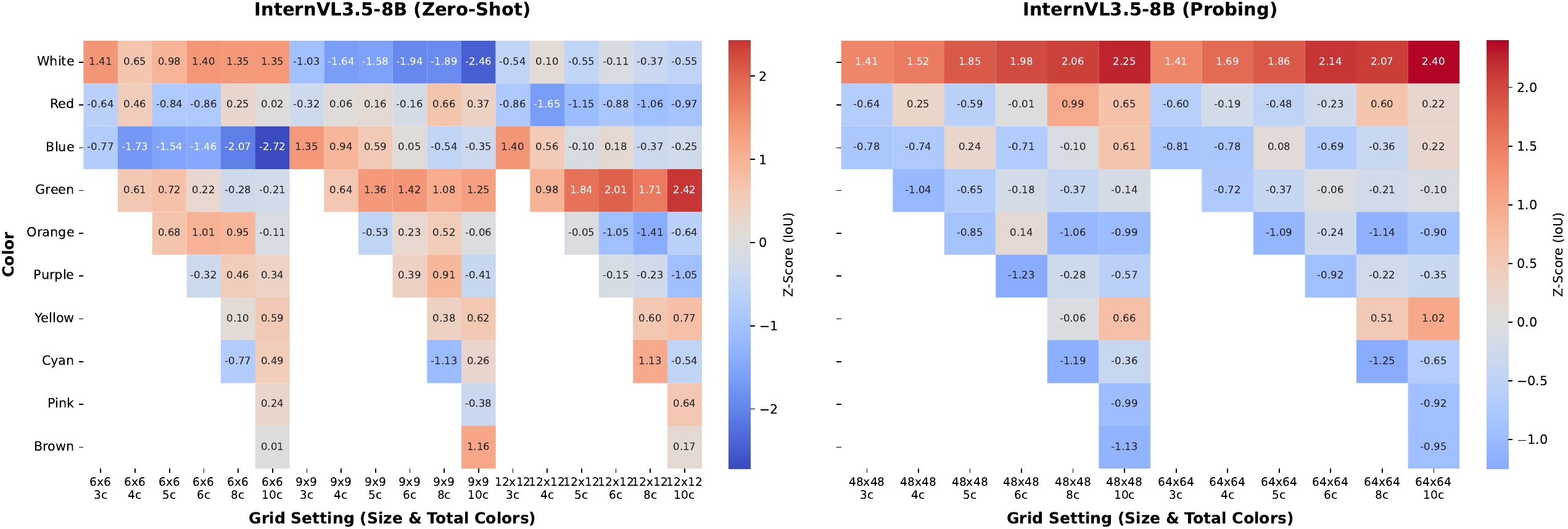} 
    \caption{Color analysis heatmap illustrating per-color localization performance in InternVL3.5-8B.}
    \label{fig:color_analysis_internvl}
\end{figure}

\textbf{Probing and Zero-Shot Favor Different Colors.}
Finally, isolating performance by specific colors reveals a notable difference between the information encoded by the vision encoder and the features prioritized by the LLM. To rigorously evaluate spatial localization per color, we report the Intersection over Union (IoU), defined for a given color $c$ as $\text{IoU}_c = \frac{\text{TP}_c}{\text{TP}_c + \text{FP}_c + \text{FN}_c}$. By jointly accounting for true positives (TP), false positives (FP), and false negatives (FN), this metric strictly penalizes degenerate spatial mappings, such as only predicting a single color, which would otherwise artificially inflate the accuracy for that color.

When probing the isolated vision encoder, \textit{White} is consistently the easiest color to detect, followed by \textit{Yellow}, dominating the IoU metrics across all grid sizes. Yet, during zero-shot generation, the end-to-end VLM shifts its focus away from \textit{White} once the grid size exceeds the model's capacity for near-perfect transcription. At $9 \times 9$ and $12 \times 12$ densities, the zero-shot accuracy for \textit{White} plummets, and the VLM instead achieves its highest scores on \textit{Green} and \textit{Yellow}. This behavior aligns with the semantic priors established during multimodal instruction tuning: in some of the training tasks that are highly structured (such as documents, charts, and GUIs), \textit{White} overwhelmingly functions as negative background space. Consequently, when the LLM's spatial tracking becomes overwhelmed by the grid density, it appears to default to prioritizing salient foreground colors like \textit{Green} and \textit{Yellow}, effectively discarding or ignoring the White features that the vision encoder is the best at extracting.

\section{Scaling Analysis of Proprietary Models}
\label{app:analysis_scaling}

\begin{table}[htbp]
\centering
\small
\setlength{\tabcolsep}{5pt}
\renewcommand{\arraystretch}{1.05}
\caption{Zero-shot \textit{Exact / Cell} accuracies (\%) of proprietary VLMs on G2M for grid densities $6 \times 6$ to $32 \times 32$.}
\label{tab:closed_source_scaling}
\begin{tabular}{@{}lcccc@{}}
\toprule
\textbf{Grid} & \textbf{GPT-5-mini} & \textbf{GPT-5.2} & \textbf{Gemini-3-Flash} & \textbf{Gemini-3-Pro} \\
\midrule
$6\times6$   & 97.3 / 99.9 & 100.0 / 100.0 & 99.0 / 100.0 & 100.0 / 100.0 \\
$9\times9$   & 0.0 / 65.6  & 97.3 / 100.0  & 97.3 / 100.0 & 100.0 / 100.0 \\
$12\times12$ & 0.0 / 41.5  & 94.3 / 100.0  & 69.7 / 99.2  & 100.0 / 100.0 \\
$20\times20$ & 0.0 / 33.7  & 0.0 / 64.4    & 0.0 / 87.8   & 47.3 / 84.3 \\
$32\times32$ & 0.0 / 0.0   & 0.0 / 34.1    & 0.0 / 38.5   & 0.0 / 35.0 \\
\bottomrule
\end{tabular}
\end{table}

\begin{figure}[t]
    \centering
    \includegraphics[width=\linewidth]{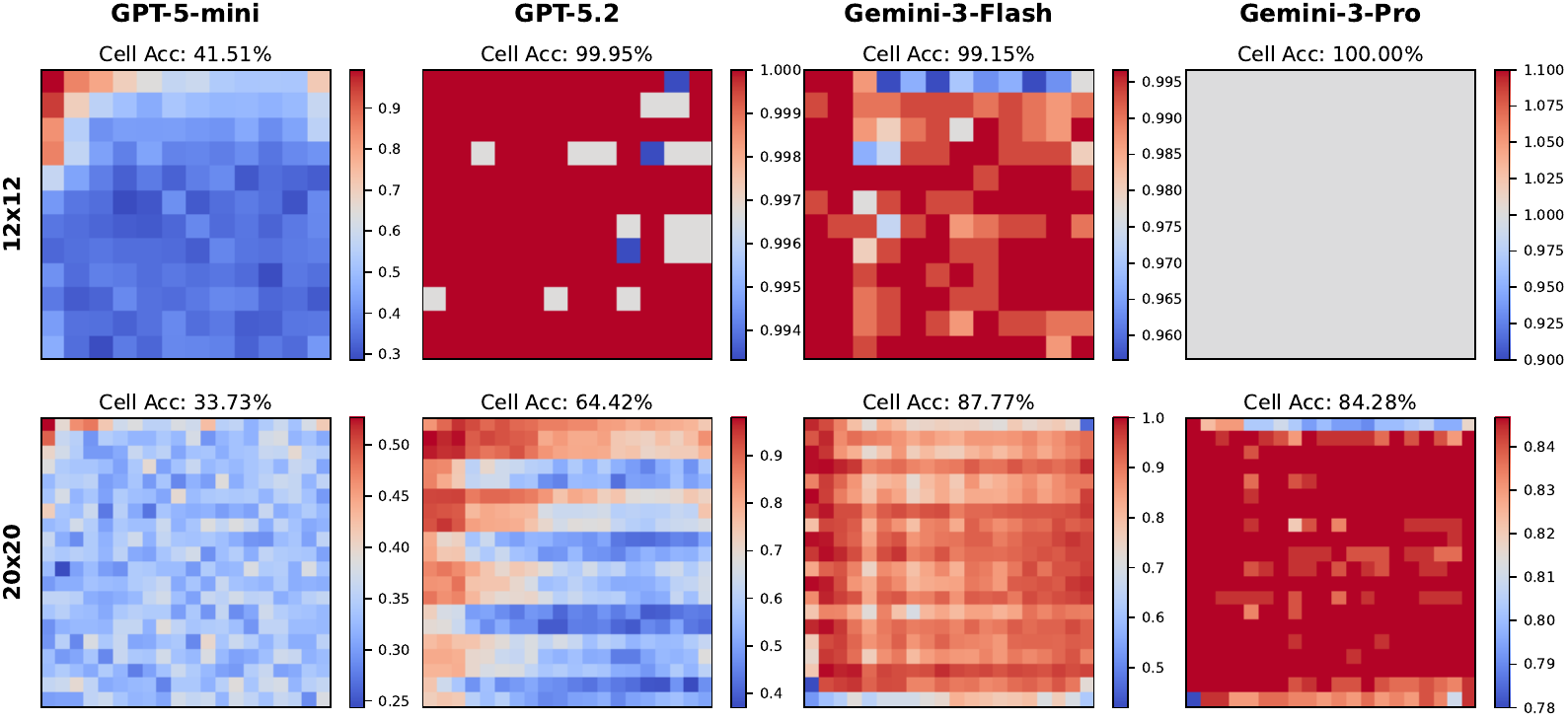} 
    \caption{Scaling on proprietary models on $12 \times 12$ and $20 \times 20$ grids.}
    \label{fig:scaling_zeroshot_closed}
\end{figure}

In addition to the open-weight models presented in Section~\ref{sec:inefficacy_scaling}, we also observed specific scaling anomalies within proprietary models.

GPT-5.2 significantly outperforms GPT-5-mini, maintaining near-perfect Cell Accuracy (99.95\%) up to $12 \times 12$ grids, but its performance heavily degrades to 64.42\% at $20 \times 20$. Interestingly, spatial heatmaps in Figure~\ref{fig:scaling_zeroshot_closed} reveal that GPT-5.2 exhibits distinct horizontal strides of failure at $20 \times 20$. We hypothesize these horizontal strides of failure at $20 \times 20$ densities arise from the model's visual tokenization process: reading the image left-to-right, row-by-row, causes a momentary lapse in spatial awareness whenever the autoregressive sequence wraps from the end of one row back to the beginning of the next. This repeated context fragmentation leads to horizontal error stripes in the final output.

More surprisingly, the Gemini 3 family exhibits a scaling inversion at extreme densities. While Gemini-3-Pro demonstrates better stability on mid-tier grids (maintaining 100\% Cell Accuracy up to $12 \times 12$), it underperforms the smaller Gemini-3-Flash model on ultra-dense grids. At $20 \times 20$, the Pro model's accuracy degrades evenly across the entire grid surface to 84.28\%, whereas the Flash model exhibits higher spatial variance but maintains a higher overall accuracy of 87.77\%. By $32 \times 32$, the Pro model collapses to 34.99\% Cell Accuracy, while the Flash model retains 38.46\%. We hypothesize that the performance inversion—where the smaller Flash model outperforms the Pro model on ultra-dense grids—is driven by the regularization effects inherent to model compression. The Flash variant is designed for efficiency, typically achieved through distillation from a larger flagship model \citep{gemini2024}. We hypothesize that because the heavily parameterized Pro model functions as a generalized reasoning engine, it may be attempting to apply complex, semantic heuristics to the visual input, which can fail at extreme densities. In contrast, the restricted capacity and regularized nature of the smaller model strip away these conflicting abstractions, forcing it to rely strictly on the explicit in-context visual data \citep{gemma2025}. For pure spatial transcription tasks like G2M, this enforced literalism enables the smaller model to outperform its general-purpose counterpart.

\section{Grid2Matrix Evaluation Details}
\label{app:evaluation_details}

To rigorously investigate the ``Digital Agnosia'' hypothesis, our evaluation isolates the perceptual capacity of the VE from the downstream language model. We achieve this via a diagnostic spatial probe on the frozen VE, as well as an end-to-end zero-shot inference task on the full VLM.

\subsection{Data Generation and Evaluation Metrics}

\paragraph{Dataset Construction}
For probing, we use $8,000$ training samples, $2,000$ validation samples, and $10,000$ testing samples. For zero-shot inference, we use $10,000$ testing samples for open-weight models, and $300$ for proprietary models. Grid complexities range from $20 \times 20$ up to ultra-dense $64 \times 64$ layouts, with cell colors sampled uniformly from a defined dictionary. Standard input images are generated at a $512 \times 512$ resolution. 

\paragraph{Primary Metrics}
We report two primary metrics across both experimental setups:
\begin{itemize}
    \item \textbf{Exact Match (Grid-level):} A binary metric requiring zero errors across the entire matrix.
    \item \textbf{Cell Accuracy (Pixel-level):} The percentage of individual cells correctly predicted or transcribed within a given grid. 
\end{itemize}

\paragraph{Spatial Heatmaps}
To move beyond aggregate metrics, we accumulate the pixel-level correctness across the entire test set to construct continuous spatial heatmaps. For both the zero-shot generation and the spatial probes, this allows us to precisely quantify the model's perception abilities at different locations. It highlights whether failures are uniformly distributed, tied to architectural patch boundaries, clustered in specific regions, or decay with some universal patterns.

\subsection{Vision Encoder Probing Protocol}
\label{app:probing_details}

By freezing the VLM backbone and training a shallow convolutional spatial probe on the raw visual representations, we measure the exact structural information preserved by the vision tower.

\paragraph{Feature Extraction and Architecture}
We extract features from two representative open-weight architectures: Qwen3-VL-8B and InternVL3.5-8B. To ensure a fair comparison, both models process the standard $512 \times 512$ input image into an equivalent number of visual tokens before alignment.
\begin{itemize}
    \item \textbf{Qwen3-VL-8B-Instruct:} Relies on a SigLIP2-SO based vision encoder, dividing the input into $16 \times 16$ pixel patches. This yields $32 \times 32 = 1024$ visual tokens, with a hidden dimension of $1152$. In the standard forward pass, a Vision-Language Merger module compresses $2 \times 2$ neighboring features into a single token. We extract the features immediately \textit{before} this Merger module.
    \item \textbf{InternVL3.5-8B:} Reshapes the input into $448 \times 448$ pixels, which it divides into $14 \times 14$ pixel patches. This yields exactly $32 \times 32 = 1024$ visual tokens (plus one CLS token), with a hidden dimension of $1024$. We extract the final hidden state of its vision encoder, \textit{prior} to the ``Pixel Shuffle'' downsampling layer.
\end{itemize}

\paragraph{Spatial Probe Design}
Because the task requires precise pixel-to-coordinate localization, flattening all image features would cause the input dimension to explode. Instead, we use a \textbf{Spatial Probe} ($f_{\text{probe}}$) designed to independently predict the color class for every individual cell in the $N \times N$ grid.

Given an extracted sequence of visual features $X \in \mathbb{R}^{B \times L \times D}$, where $L$ is the sequence length and $D$ is the hidden dimension:
\begin{enumerate}
    \item \textbf{Reshape:} We reshape the sequence into a 2D spatial feature map of size $(B, D, h, w)$, where $h=w=\sqrt{L}$.
    \item \textbf{Interpolate:} Because the native feature map resolution ($h \times w$) cannot natively match every target grid density ($N \times N$), we apply bilinear interpolation to resize the feature map to the exact $N \times N$ dimensions.
    \item \textbf{Classification:} We project the features using a $1 \times 1$ Convolutional Head. This is mathematically equivalent to training a shared linear classifier that sweeps across and operates on every grid cell independently.
    $$Y=\text{Softmax}(\text{Conv}_{1\times1}(\text{GELU}(\text{BN}(\text{Conv}_{1\times1}(X_{\text{resized}})))))$$
\end{enumerate}
The probe head projects the input dimension to a hidden dimension of $512$, followed by Batch Normalization, a GELU activation, and a final projection to the number of target color classes ($C$).

\paragraph{Training Protocol}
The probes are trained using the standard cross-entropy loss for multi-class classification. We optimize using AdamW with a learning rate of $1 \times 10^{-2}$ and a weight decay of $1 \times 10^{-4}$. The models are trained using a batch size of $32$ in mixed precision (bfloat16/float16). We employ a cosine learning rate schedule with a 5\% warmup period, training for a maximum of $5,000$ iterations. Training takes approximately 1 hour on a single NVIDIA RTX A6000 GPU for an 8B parameter model.

\subsection{Zero-Shot Inference Protocol}
\label{app:zeroshot_details}

Zero-shot evaluation tests the model's end-to-end capability to perceive the dense visual grid and accurately articulate its spatial information as a textual matrix. Evaluating an 8B parameter model on a single RTX A6000 GPU takes approximately 20 minutes for a $4 \times 4$ grid, scaling up to 3 hours for a $12 \times 12$ grid.

\paragraph{Prompting and Generation Constraints}
To isolate pure spatial perception from logical deduction, we deliberately avoid Chain-of-Thought (CoT) prompting. The model receives a concise instruction to transcribe the image into a Python list of lists, accompanied by a dynamic color-to-integer dictionary (e.g., \texttt{\{White: 0, Red: 1, Blue: 2\}}). The exact prompt template used across all evaluations is as follows:

\begin{quote}
\textbf{System/User Instruction:} \\
You are a precise grid serialization engine. \\
Task: Transcribe the $H \times W$ pixel grid from the image into a numerical matrix. \\
Color Mapping: \{White: 0, Red: 1, Blue: 2, \dots\} \\
\\
Instructions: \\
1. Scan the grid row by row, from top to bottom. \\
2. For each row, map every cell using the color mapping. \\
3. Ensure the output has exactly $H$ rows and $W$ columns. \\
\\
Output Format: \\
Return ONLY a Python list of lists (e.g., [[0, 1], [2, 0]]). Do not use markdown, code blocks, or explanations.
\end{quote}

To prevent prompt instability, the color mapping dictionary is strictly sorted by its integer values before being injected into the prompt template. We enforce fully deterministic generation using greedy decoding (temperature of $0$) to eliminate sampling noise. Inference is conducted in \texttt{bfloat16} or \texttt{float16} to perfectly match the native training precision of the evaluated models. To prevent premature cutoffs while avoiding computationally expensive infinite loops, we dynamically calculate the maximum token limit based on the target grid dimensions: 
$$N_{\text{max\_tokens}}=\min((H \times W \times 4) + (H \times 20) + 50, 2048)$$

\paragraph{Robust Output Parsing}
Frontier VLMs frequently deviate from strict formatting constraints (e.g., adding markdown blocks or conversational filler). To ensure we evaluate spatial accuracy rather than instruction-following compliance, we process the raw text through a cascading parser:
\begin{enumerate}
    \item \textbf{Strict Evaluation:} We first attempt to directly parse the output stream as a valid Python array using \texttt{ast.literal\_eval}.
    \item \textbf{Row-wise Regex:} If standard parsing fails, we apply regular expressions to catch row-by-row structural patterns (e.g., \texttt{ROW1=[...]}), a format some models spontaneously adopt when overwhelmed.
    \item \textbf{Fallback Flattening:} As a final contingency, we extract all integers from the text response. If the total integer count perfectly matches the expected $H \times W$ cells, we sequentially reshape this 1D array into the target 2D grid dimensions.
\end{enumerate}
If all three extraction methods fail, the sample is recorded as a complete parse error and yields a 0\% Cell Accuracy for that grid.

\section{More Result Details}

\begin{table}[ht]
\centering
\small
\setlength{\tabcolsep}{5pt}
\renewcommand{\arraystretch}{1.05}
\caption{Zero-shot \textit{Exact / Cell} accuracies (\%) of open-weight models on 3-color grids.}
\label{tab:open_source_zeroshot}
\begin{tabular}{@{}lcc@{}}
\toprule
\textbf{Grid} & \textbf{InternVL3.5-8B} & \textbf{Qwen3-VL-8B-Instruct} \\
\midrule
$2 \times 2$   & 100.0 / 100.0 & 100.0 / 100.0 \\
$3 \times 3$   & 100.0 / 100.0 & 66.0 / 99.8 \\
$4 \times 4$   & 100.0 / 100.0 & 0.0 / 96.6 \\
$6 \times 6$   & 62.0 / 98.1   & 0.0 / 71.2 \\
$9 \times 9$   & 0.0 / 64.8    & 0.0 / 40.0 \\
$12 \times 12$ & 0.0 / 35.6    & 0.0 / 35.0 \\
\bottomrule
\end{tabular}
\end{table}

\begin{table}[ht]
\centering
\small
\setlength{\tabcolsep}{5pt}
\renewcommand{\arraystretch}{1.05}
\caption{Spatial Probe \textit{Exact / Cell} accuracies (\%) on high-density 3-color grids.}
\label{tab:ve_probe_results}
\begin{tabular}{@{}lcc@{}}
\toprule
\textbf{Grid} & \textbf{InternVL3.5-8B} & \textbf{Qwen3-VL-8B-Instruct} \\
\midrule
$20 \times 20$ & 100.0 / 100.0 & 75.1 / 99.9 \\
$32 \times 32$ & 100.0 / 100.0 & 96.2 / 100.0 \\
$48 \times 48$ & 0.0 / 98.8    & 0.0 / 72.3 \\
$64 \times 64$ & 0.0 / 69.3    & 0.0 / 54.8 \\
\bottomrule
\end{tabular}
\end{table}

\begin{figure}[t]
    \centering
    \includegraphics[width=\linewidth]{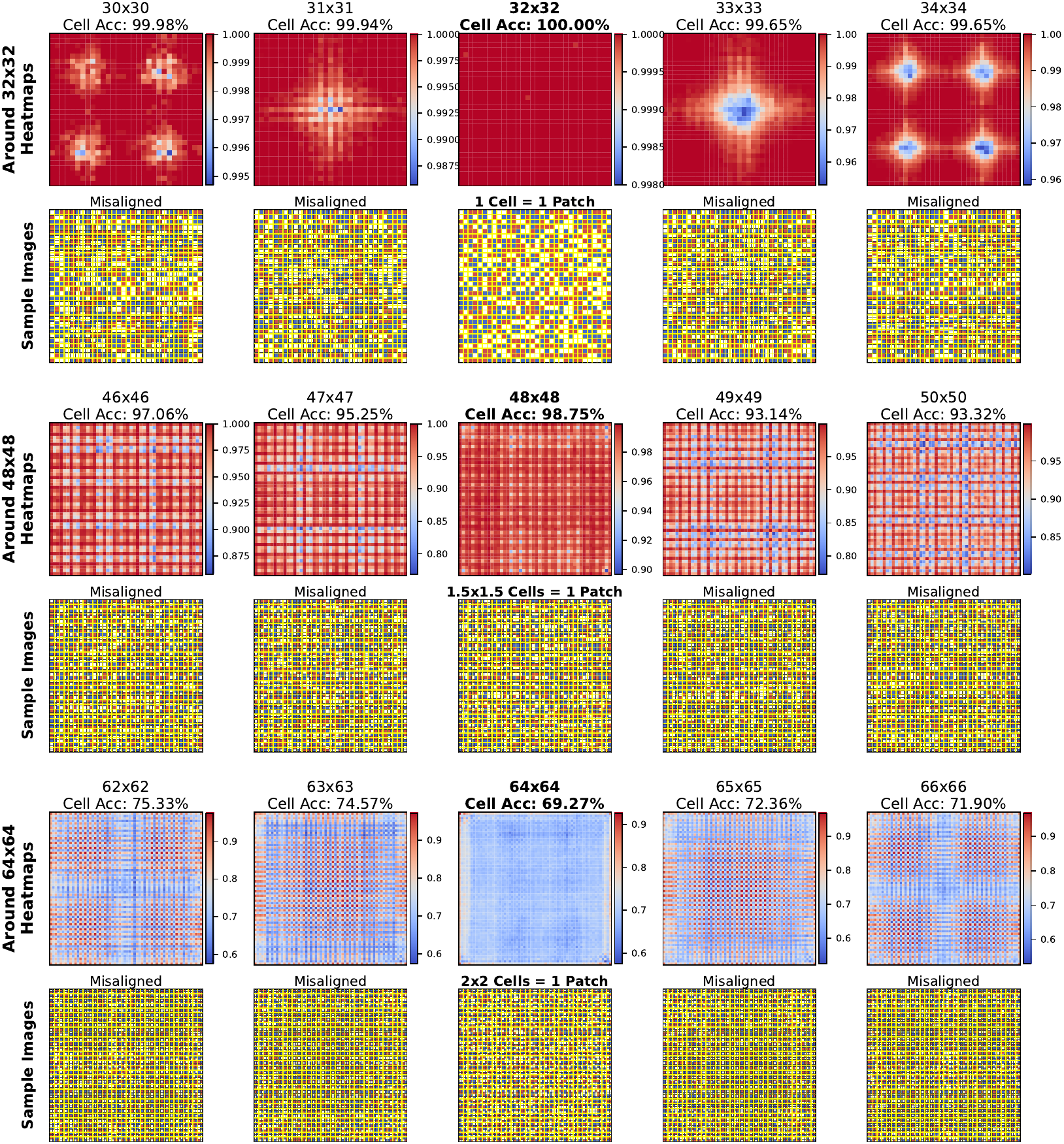} 
    \caption{Grid-patch interaction patterns in InternVL3.5-8B. We present error heatmaps for grid sizes around $32 \times 32$, $48 \times 48$, and $64 \times 64$. Below each heatmap, yellow lines illustrate how the vision encoder's patch boundaries overlay the original image. Because the fixed patch size does not evenly divide the grid cells, the boundaries progressively misalign across the image. As detailed in Section~\ref{sec:patch_grid_geometry}, the cell that occupies the majority of a patch dictates the feature representation and achieves higher accuracy. This gradual shift in patch-to-cell overlap produces the distinct macroscopic interference patterns visible in the error heatmaps.}
    \label{fig:full-alias-internvl}
\end{figure}

\begin{figure}[t]
    \centering
    \includegraphics[width=\linewidth]{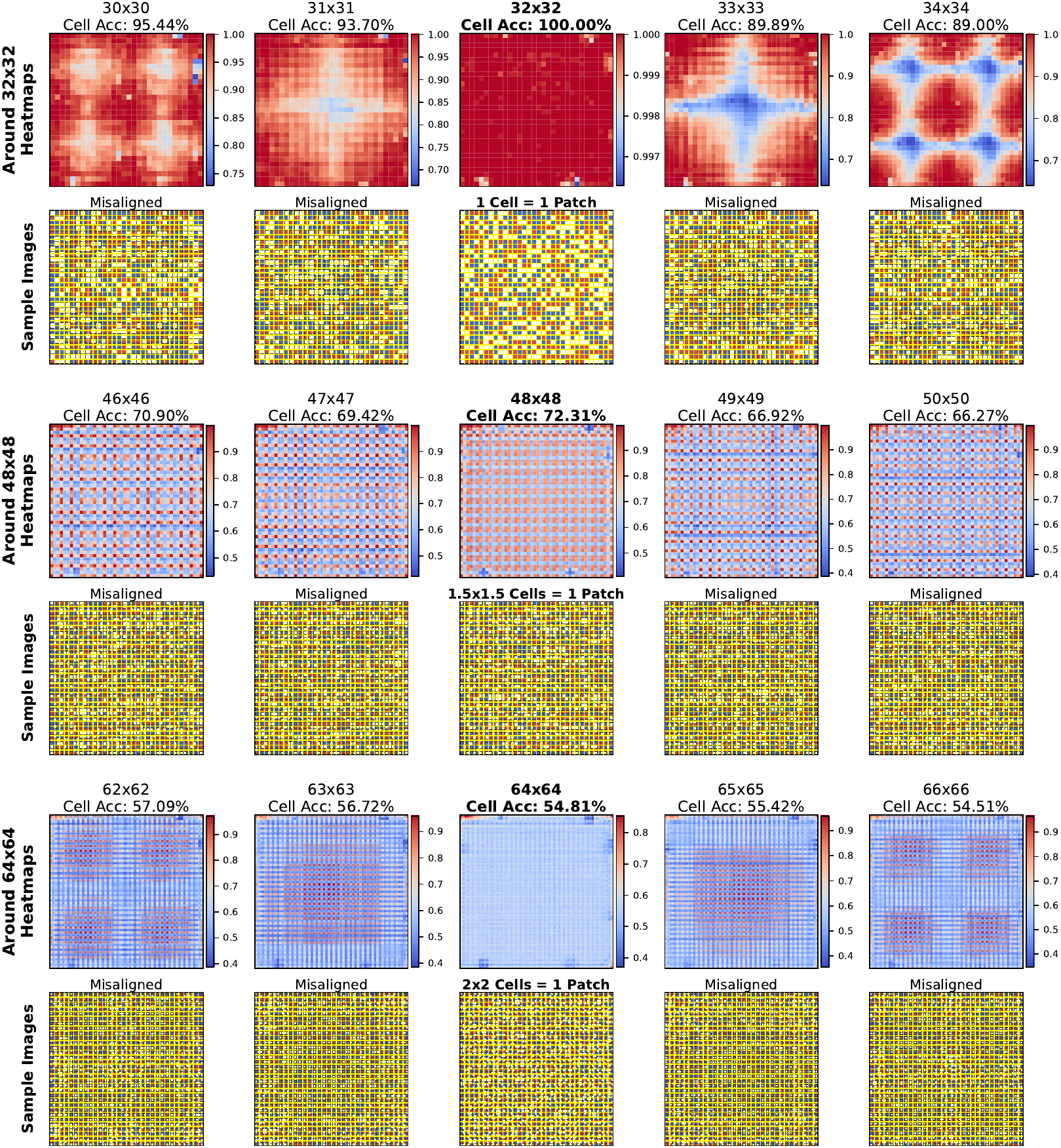} 
    \caption{Grid-patch interaction patterns in Qwen3-VL-8B-Instruct. We present error heatmaps for grid sizes around $32 \times 32$, $48 \times 48$, and $64 \times 64$. Below each heatmap, yellow lines illustrate how the vision encoder's patch boundaries overlay the original image. Because the fixed patch size does not evenly divide the grid cells, the boundaries progressively misalign across the image. As detailed in Section~\ref{sec:patch_grid_geometry}, the cell that occupies the majority of a patch dictates the feature representation and achieves higher accuracy. This gradual shift in patch-to-cell overlap produces the distinct macroscopic interference patterns visible in the error heatmaps. The impact of grid-patch interactions appears to be more pronounced compared to InternVL3.5-8B, which is especially visible in the first row (around $32 \times 32$).}
    \label{fig:full-alias-qwen}
\end{figure}

\begin{figure}[t]
    \centering
    \includegraphics[width=\linewidth]{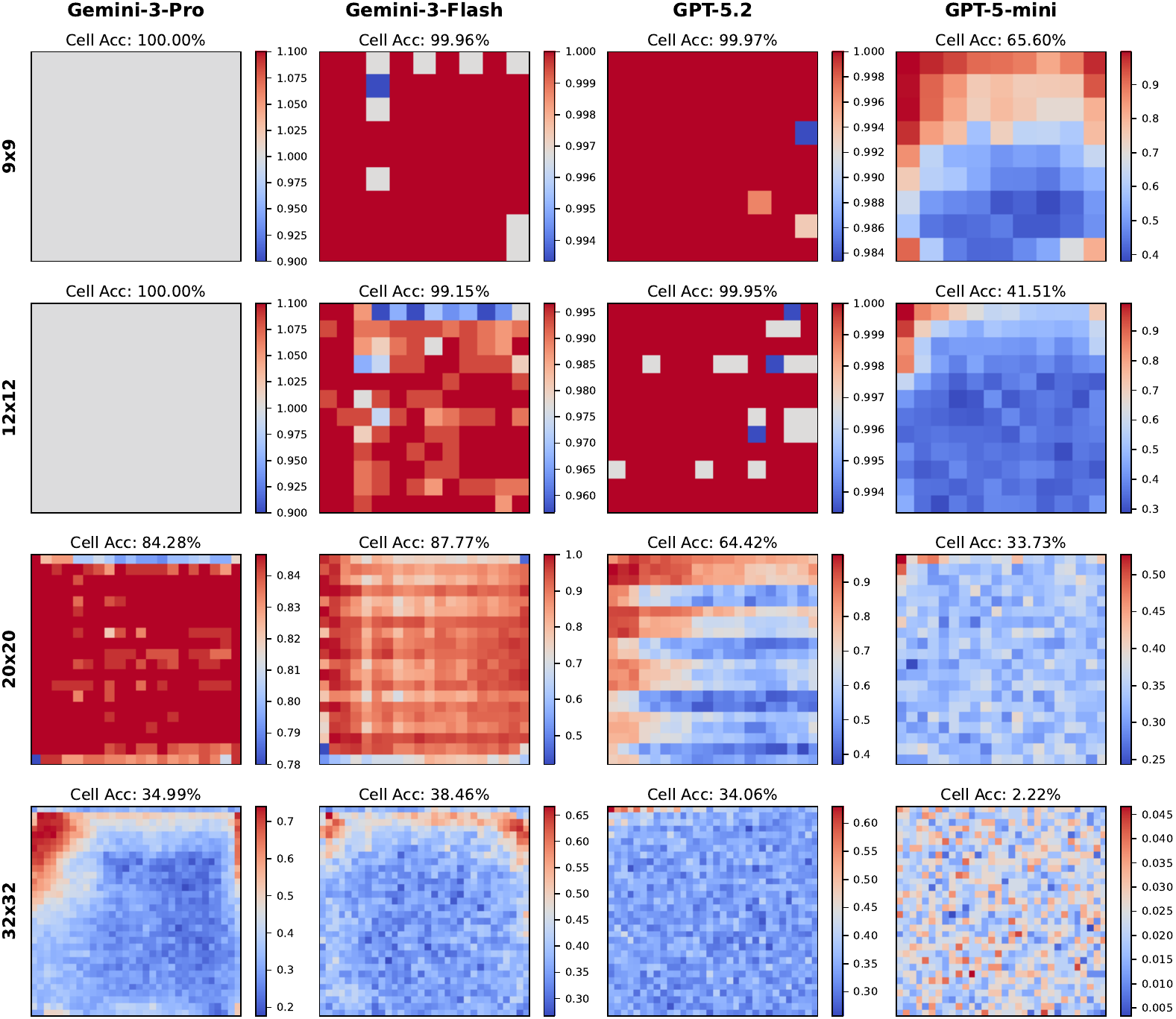} 
    \caption{Zero-shot spatial error heatmaps for all proprietary models, a full version of Figure~\ref{fig:scaling_zeroshot_open} from Section~\ref{sec:inefficacy_scaling}.}
    \label{fig:heatmap_closed_full}
\end{figure}

\begin{figure}[t]
    \centering
    \includegraphics[width=\linewidth]{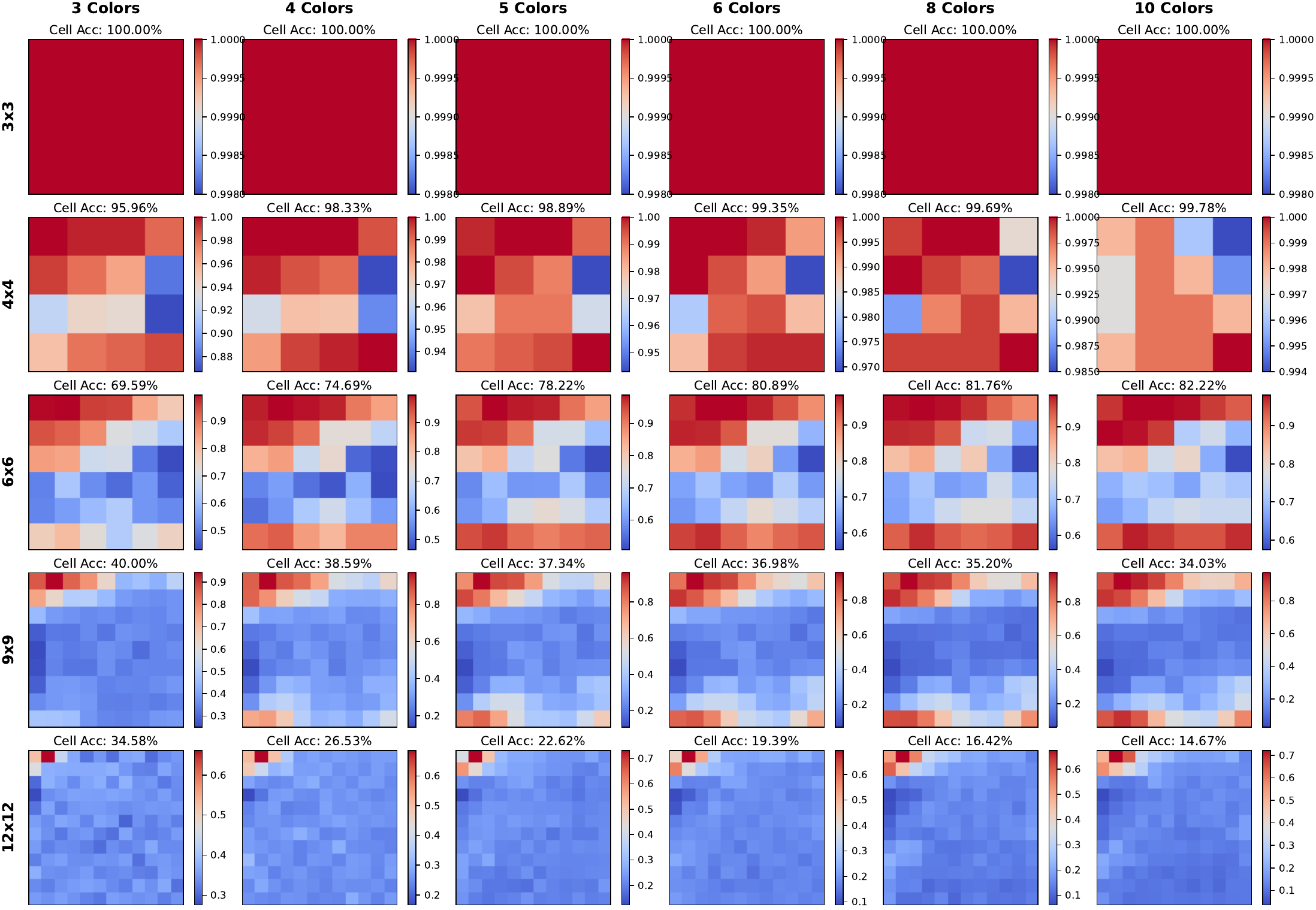} 
    \caption{Zero-shot spatial error heatmaps for Qwen3-VL-8B-Instruct across varying grid sizes and color complexities. The autoregressive degradation pattern remains as the color vocabulary increases (see Appendix~\ref{sec:color_ablations}).}
    \label{fig:color_size_qwen_zs}
\end{figure}

\begin{figure}[t]
    \centering
    \includegraphics[width=\linewidth]{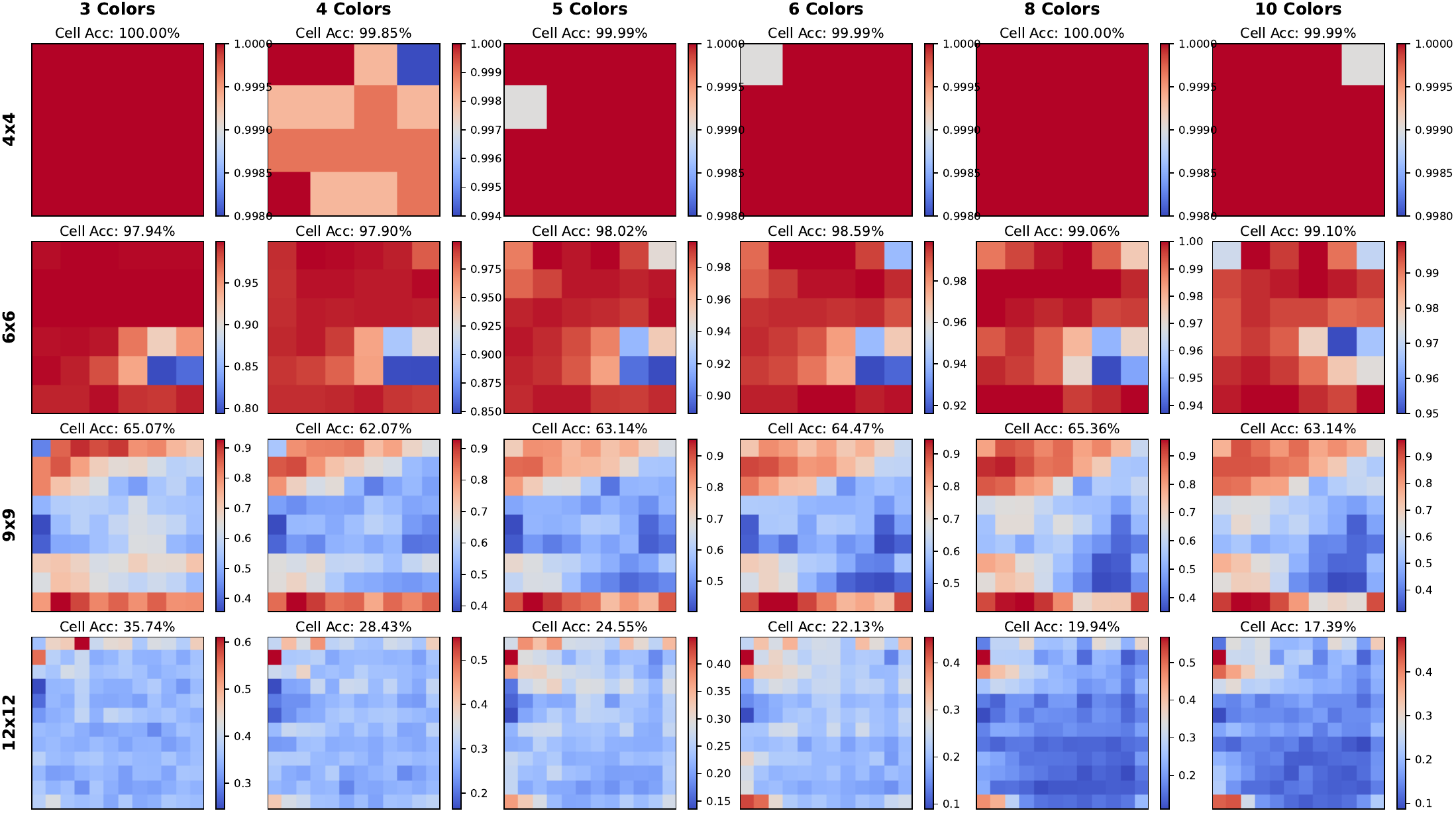} 
    \caption{Zero-shot spatial error heatmaps for InternVL3.5-8B. Similar to Qwen3-VL, the autoregressive degradation pattern remains as the color vocabulary increases (see Appendix~\ref{sec:color_ablations}).}
    \label{fig:color_size_internvl_zs}
\end{figure}

\begin{figure}[t]
    \centering
    \includegraphics[width=\linewidth]{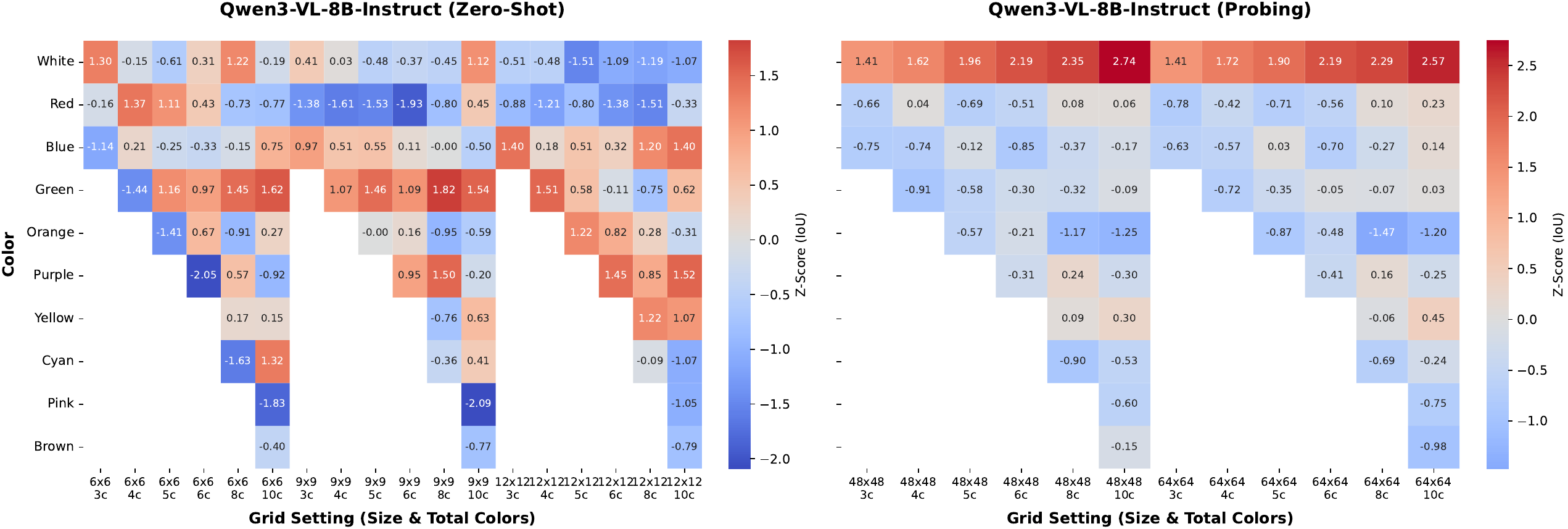} 
    \caption{Color analysis heatmap illustrating per-color localization performance in Qwen3-VL-8B-Instruct. We observe a similar pattern as for InternVL3.5-8B in Figure~\ref{fig:color_analysis_internvl} from Appendix~\ref{sec:color_ablations}.}
    \label{fig:color_analysis_qwen}
\end{figure}

\begin{figure}[t]
    \centering
    \includegraphics[width=\linewidth]{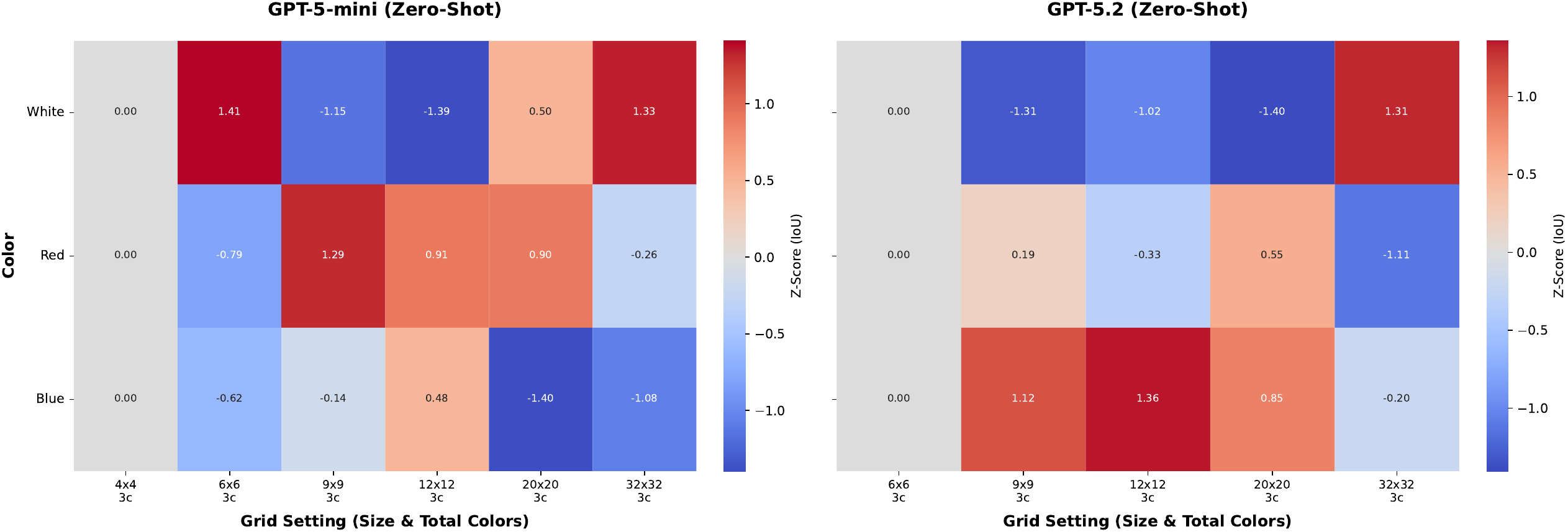} 
    \caption{Color analysis heatmap illustrating per-color localization performance in GPT-5-mini versus GPT-5.2 (see Appendix~\ref{sec:color_ablations}). We observe that both models exhibit a bias toward overpredicting blue relative to other colors. This is not immediately visible in this plot since we processed the raw accuracies with IoU. Furthermore, white represents the earliest failure mode as grid complexity increases, suffering from severe underprediction. This phenomenon is likely tied to the models' pre-training distributions, where white often serves as negative space or lacks semantic salience in natural images.}
    \label{fig:color_analysis_gpt}
\end{figure}

\begin{figure}[t]
    \centering
    \includegraphics[width=\linewidth]{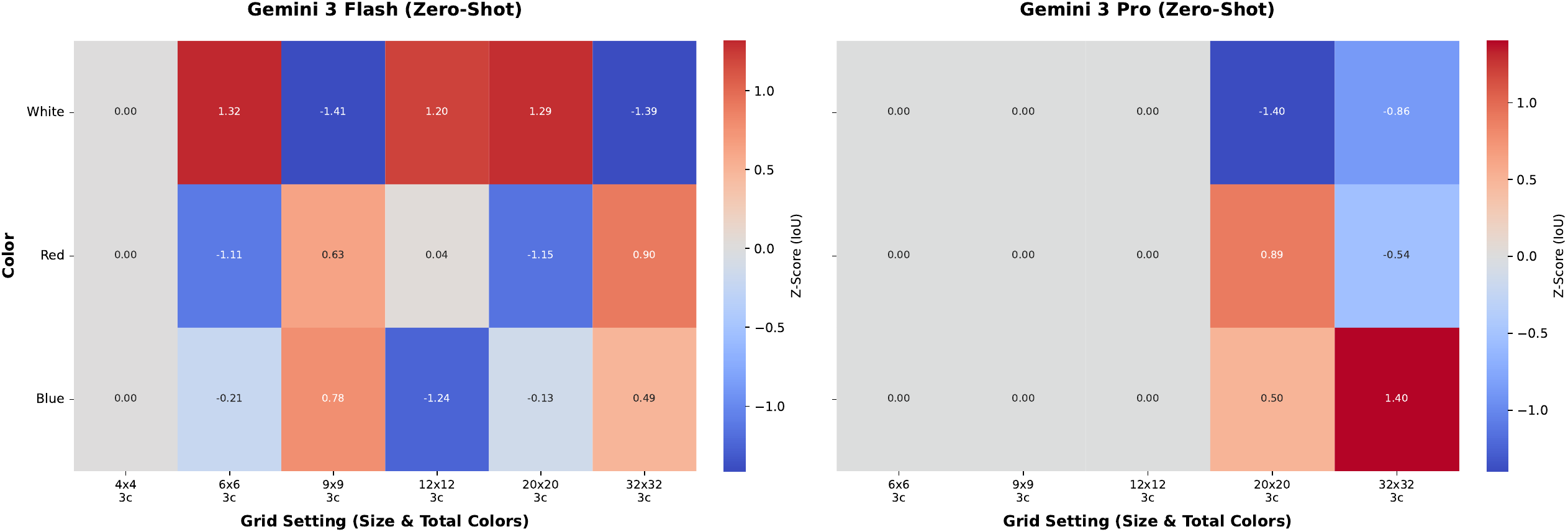} 
    \caption{Color analysis heatmap illustrating per-color localization performance in Gemini 3 Flash versus Gemini 3 Pro (see Appendix~\ref{sec:color_ablations}). Similar to the GPT-5 family, both Gemini models demonstrate a tendency to overpredict blue. Additionally, white is typically the first color to degrade in performance, becoming significantly underpredicted at larger grid sizes. We hypothesize this bias also stems from natural image distributions, where pure white patches are rarely the primary informative target.}
    \label{fig:color_analysis_gemini}
\end{figure}

\begin{figure}[t]
    \centering
    \includegraphics[width=\linewidth]{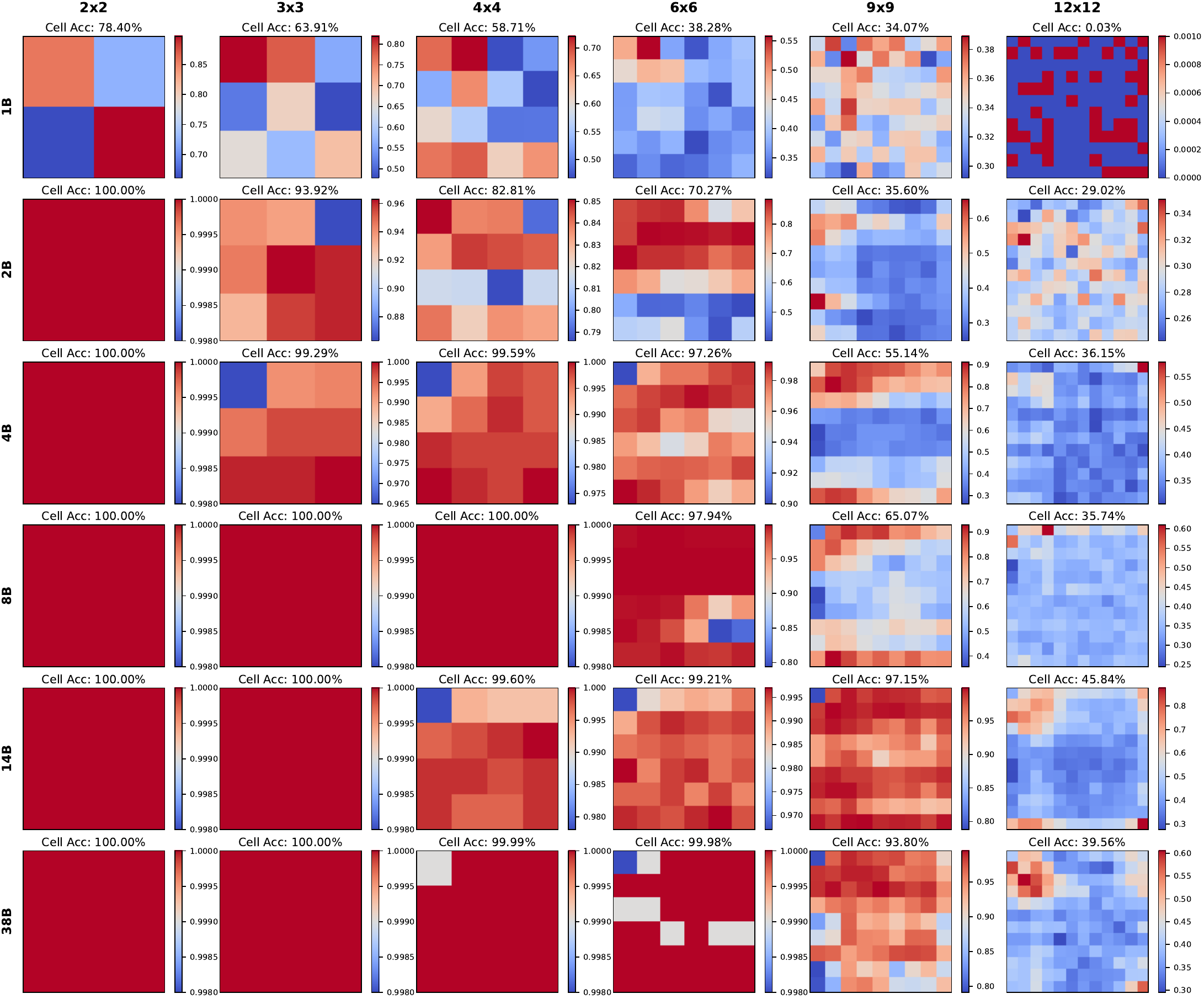} 
    \caption{Zero-shot spatial error heatmaps for the InternVL3.5 family across varying model and grid sizes. While scaling the LLM parameter count generally improves spatial retention on less dense grids, we observe a distinct spatial degradation when scaling from 14B to the 38B model on the $12 \times 12$ grid, which we hypothesize is due to the projection bandwidth bottleneck as discussed in Section~\ref{sec:inefficacy_scaling}.}
    \label{fig:grid-size-internvl}
\end{figure}

\begin{figure}[t]
    \centering
    \includegraphics[width=\linewidth]{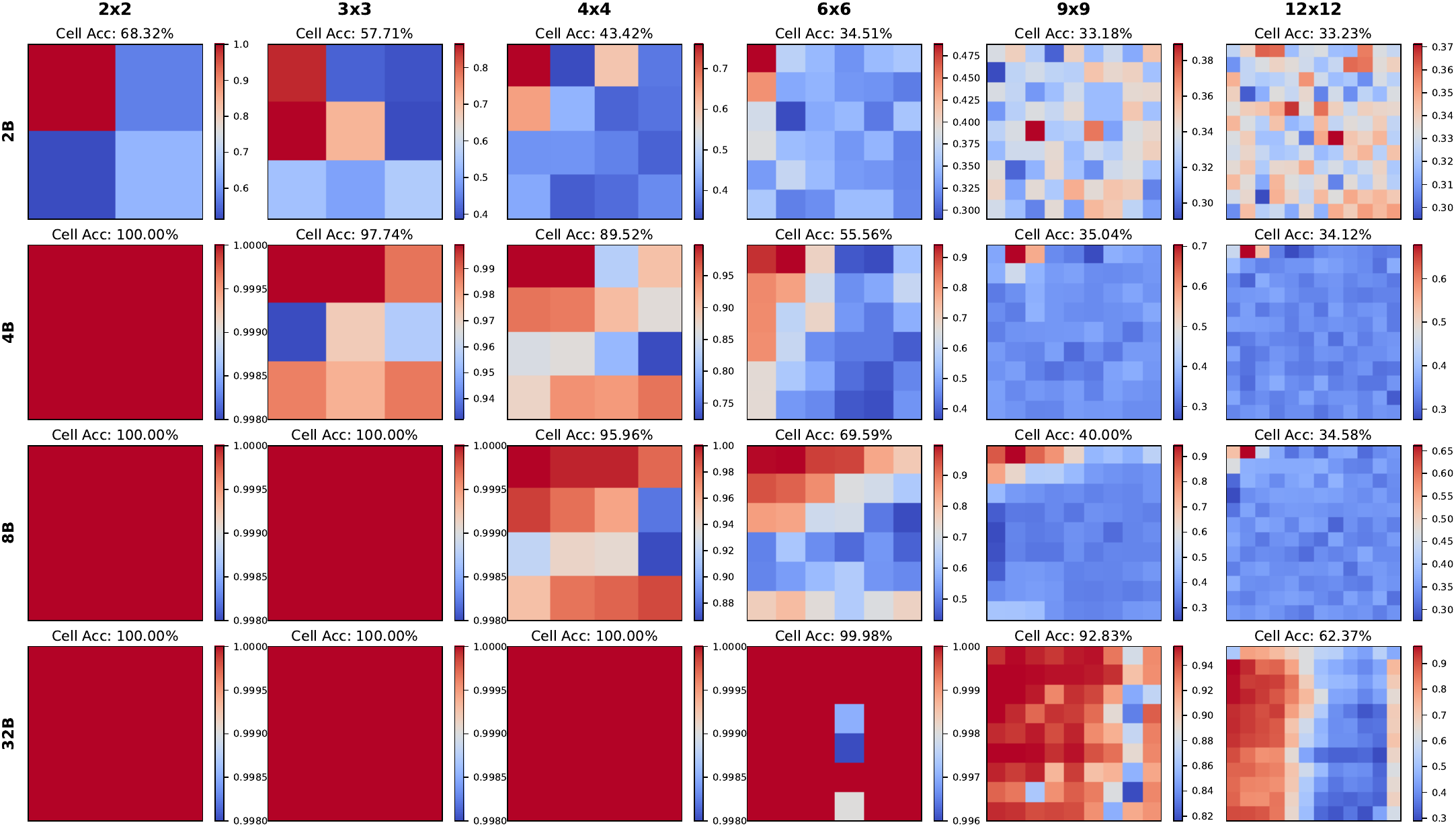} 
    \caption{Zero-shot spatial error heatmaps for the Qwen3-VL family. In contrast to InternVL, Qwen exhibits strictly monotonic improvement with end-to-end model scale. Notably, at 32B, the model successfully maintains spatial fidelity across almost the entire left half of the $12 \times 12$ grid.}
    \label{fig:grid-size-qwen}
\end{figure}

\begin{figure}[t]
    \centering
    \includegraphics[width=\linewidth]{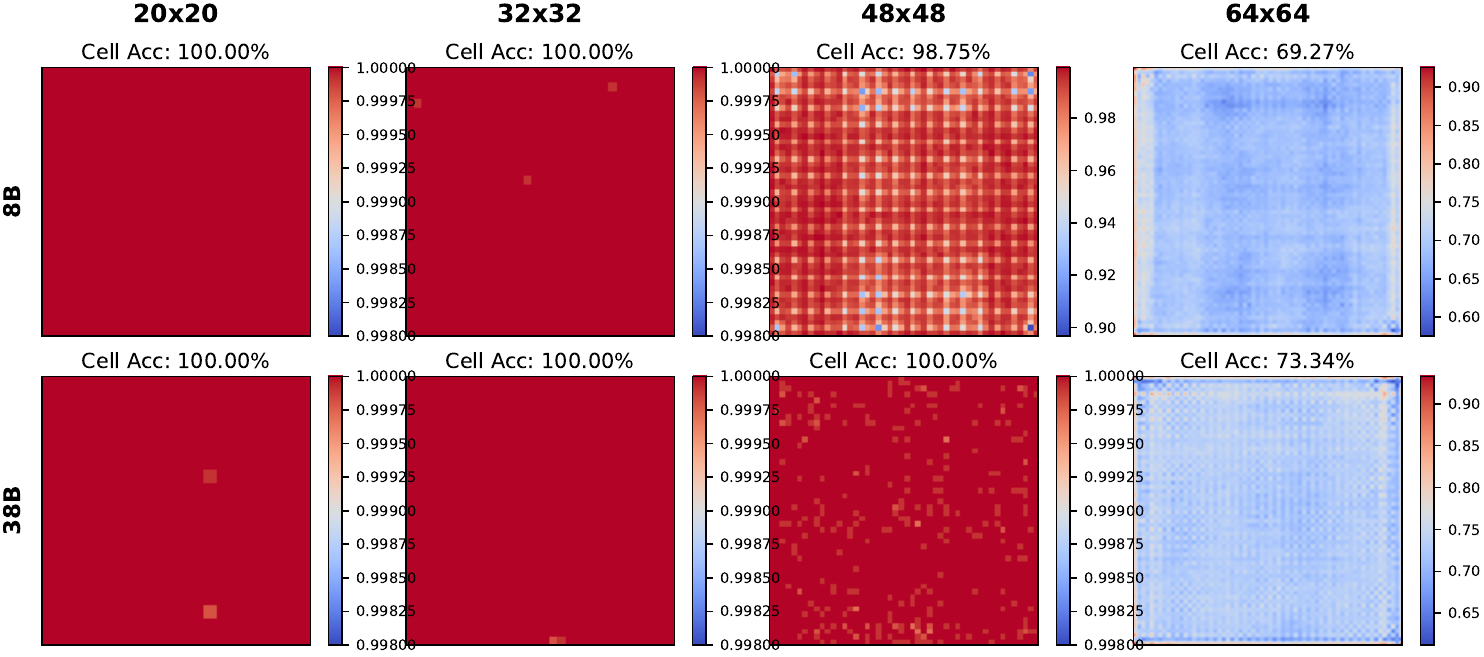} 
    \caption{VE spatial probing heatmaps for the InternVL3.5 family across varying grid sizes. The 8B model uses the InternViT-300M encoder, while the 38B model upgrades to the massive InternViT-6B. As detailed in Section~\ref{app:analysis_scaling}, scaling the VE significantly enhances the spatial fidelity captured in the raw visual representations.}
    \label{fig:grid-size-internvl-probe}
\end{figure}

\begin{figure}[t]
    \centering
    \includegraphics[width=\linewidth]{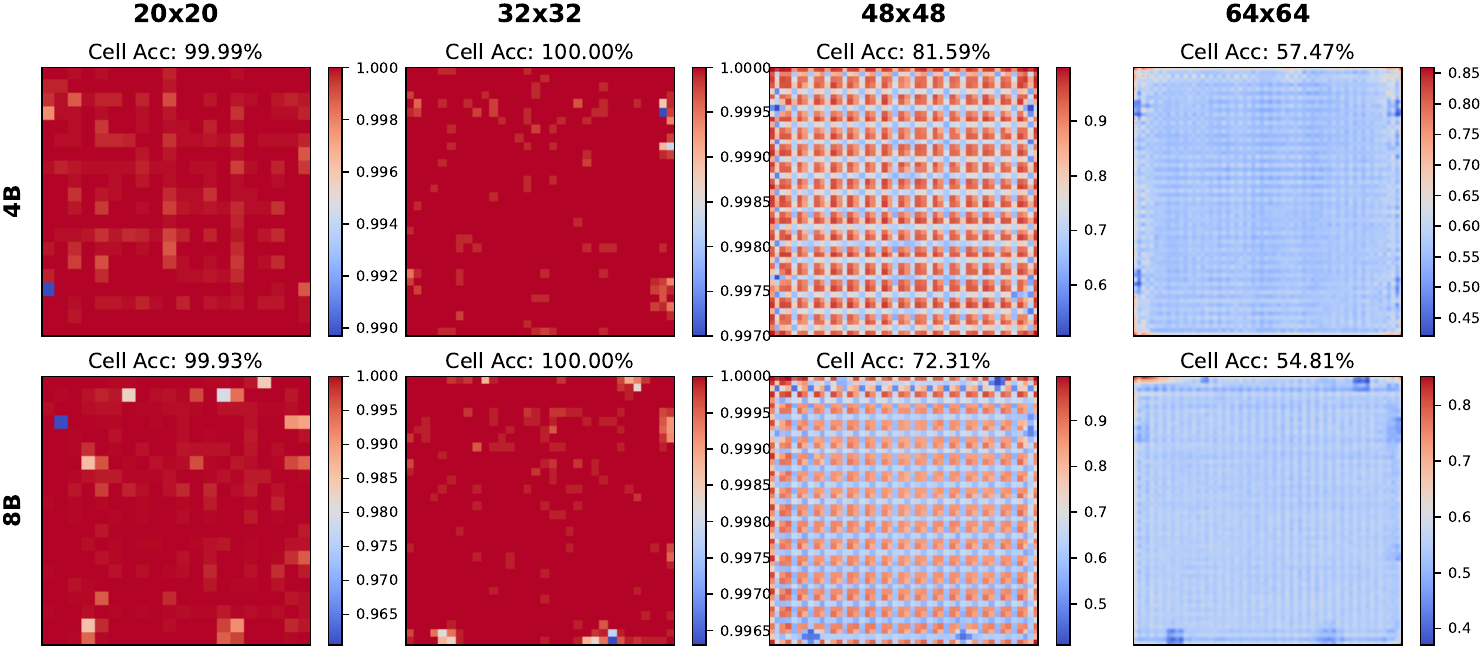} 
    \caption{VE spatial probing heatmaps for the Qwen3-VL family. The 4B model uses the SigLIP2-Large-300M encoder, whereas the 8B model uses the SigLIP2-SO-400M. Contrary to conventional scaling expectations, the larger VE actually exhibits a slight degradation in spatial fidelity compared to its smaller counterpart (discussed further in Section~\ref{app:analysis_scaling}).}
    \label{fig:grid-size-qwen-probe}
\end{figure}

\begin{figure}[t]
    \centering
    \includegraphics[width=\linewidth]{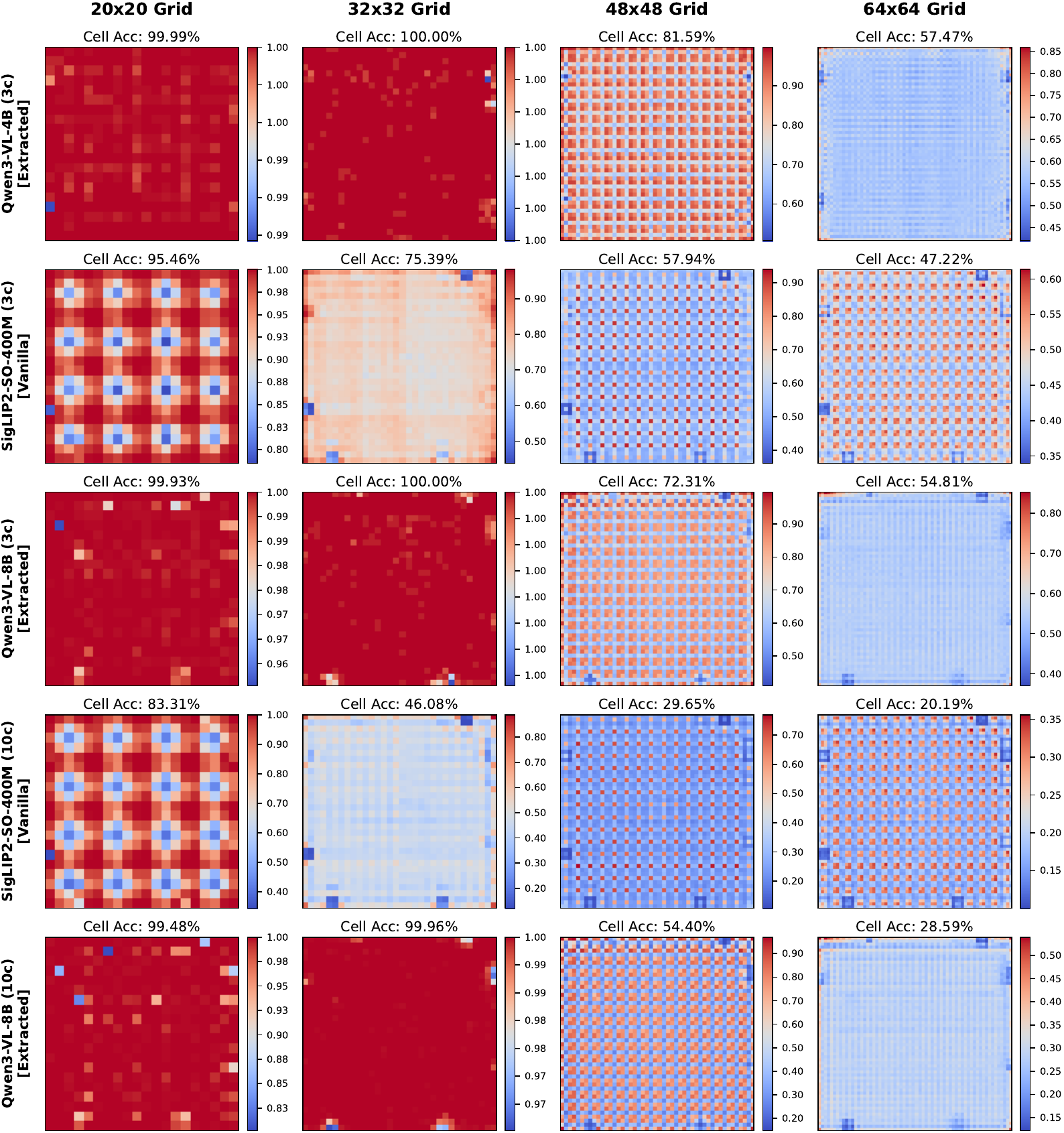} 
    \caption{VE spatial probing heatmaps for the Qwen3-VL family, comparing the isolated pre-trained vision encoder against the aligned vision encoder extracted from the full VLM. The structural blind spots remain remarkably consistent across both settings, suggesting that multimodal alignment might not fully correct these inherent spatial vulnerabilities (discussed further in Section~\ref{sec:alignment_paradox}).}
    \label{fig:alignment-qwen-probe}
\end{figure}

\begin{figure}[t]
    \centering
    \includegraphics[width=\linewidth]{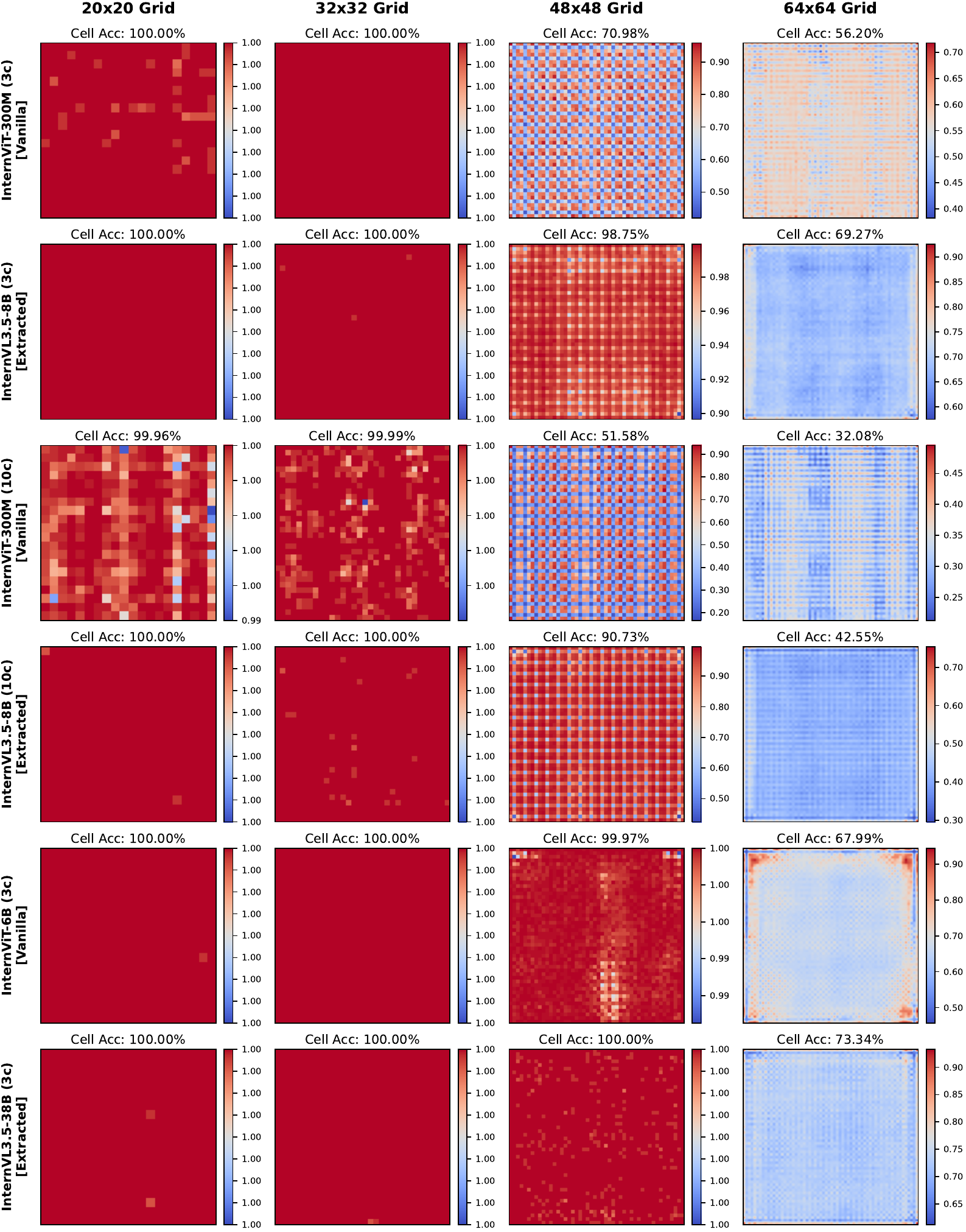} 
    \caption{VE spatial probing heatmaps for the InternVL3.5 family, comparing the pre-trained vision encoder with its aligned counterpart. In contrast to Qwen3-VL, the spatial error landscape shifts noticeably not only across different model scales but also before and after multimodal alignment, suggesting that its spatial mechanics might be more sensitive to the alignment process (discussed further in Section~\ref{sec:alignment_paradox}).}
    \label{fig:alignment-internvl-probe}
\end{figure}
\end{document}